\documentclass[10pt,twocolumn,letterpaper]{article}

\usepackage[pagenumbers]{cvpr} 

\definecolor{cvprblue}{rgb}{0.21,0.49,0.74}
\definecolor{oursrow}{HTML}{EAF3FC}
\definecolor{tablegroup}{HTML}{F0F1F3}
\usepackage[pagebackref,breaklinks,colorlinks,allcolors=cvprblue]{hyperref}
\usepackage[T1]{fontenc}
\usepackage{graphicx}
\usepackage{amsmath,amssymb}
\usepackage{booktabs,multirow}
\usepackage{xcolor}
\usepackage{colortbl}

\usepackage{times,array}
\usepackage[capitalize]{cleveref}

\def\confName{CVPR}
\def\confYear{2027}

\newcommand{\method}{ScaleCurve}

\title{ScaleCurve: From Constant Scale to Consistent Motion\\for World-Space Human Mesh Recovery}

\author{
Sanghyeon Lee \qquad
Jong Taek Lee\thanks{Corresponding author.}\\
School of Computer Science, Kyungpook National University\\
Daegu, Republic of Korea\\
{\tt\small \{hyeon1263, jongtaeklee\}@knu.ac.kr}
}

\begin{document}
\maketitle
\begin{abstract}
World-space human mesh recovery can lose spatial consistency even when camera-space body estimates are accurate. A single sequence-wide scale resolves global scale ambiguity but cannot correct local scale distortion along a reconstructed camera path. We present \method{}, a framework that estimates time-varying metric corrections from human-ground geometry. We construct scale observations by comparing predicted foot geometry with locally fitted ground planes. Joint temporal optimization downweights inconsistent observations and couples neighboring scale estimates, allowing calibration to adapt along the camera path without freely following measurement noise. We scale camera translation increments and integrate them into a calibrated trajectory. With this camera fixed, we refine the lifted body motion using 2D image evidence and pretrained 3D dynamics, while restricting global trajectory rescaling. \method{} achieves state-of-the-art performance on most evaluated metrics on both EMDB-2 and SLOPER4D. Consistent gains across three HMR backbones demonstrate its compatibility with diverse body estimates without retraining. Code will be made publicly available.
\end{abstract}
    
\section{Introduction}
\label{sec:intro}

\begin{figure}[t]
  \centering
  \includegraphics[width=\linewidth]{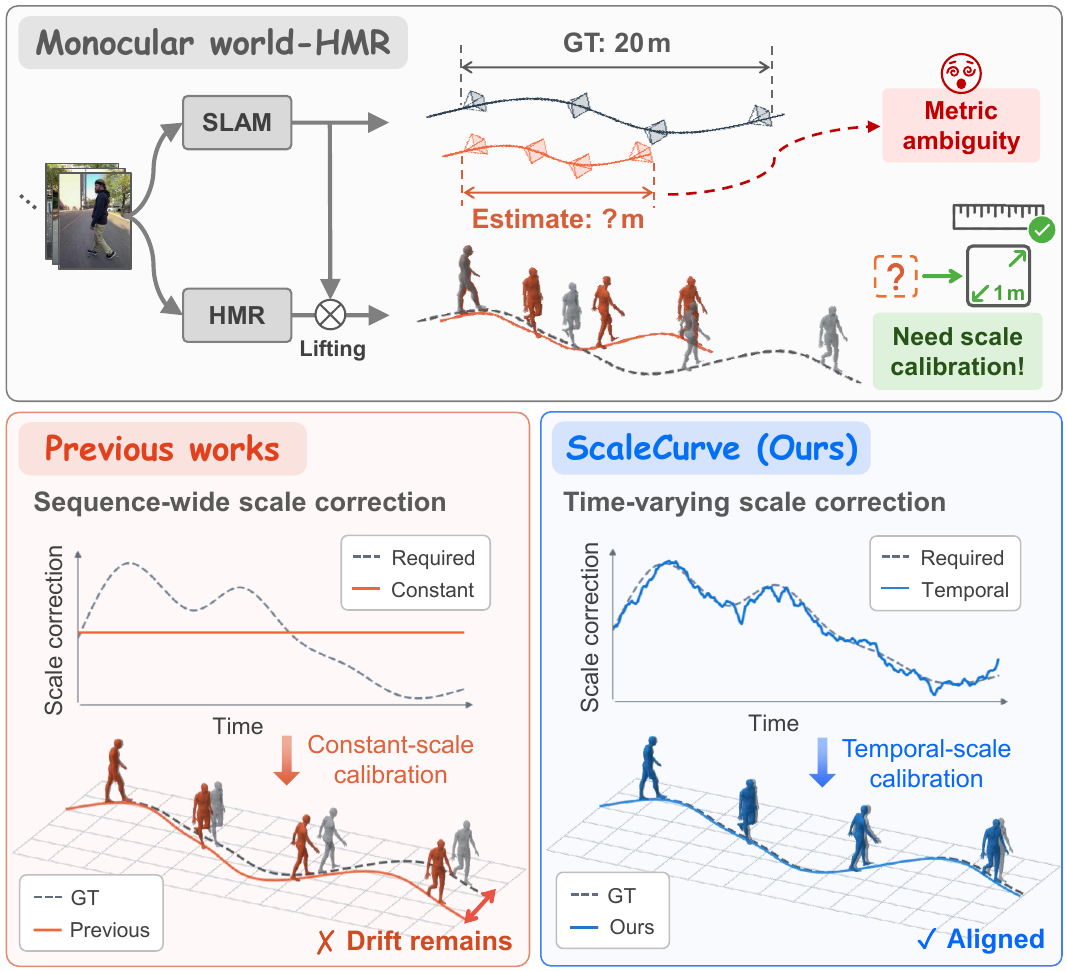}
  \caption{\textbf{Scale drift in world-space HMR.} Monocular world-space HMR inherits camera scale ambiguity. A single sequence-wide scale can correct the overall scale but leave local distortions that accumulate into human trajectory drift. \method{} instead estimates a time-varying scale correction to reduce this error.}
  \label{fig:motivation}
\end{figure}

World-space human mesh recovery (HMR) reconstructs both how a person moves and where that motion takes place in a shared scene. This spatial reference is important for reasoning about human-scene interactions and spatial relationships within the environment~\cite{dai2023sloper4d, ye2023decoupling}. Unlike camera-space recovery, world-space HMR must additionally account for camera motion. Recent approaches jointly model human and camera motion or lift camera-space body estimates using reconstructed camera trajectories~\cite{ye2023decoupling, zhao2024synergistic, shin2024wham, wang2024tram, shen2024world, wang2026duomo}. The latter strategy benefits from strong pretrained HMR and SLAM models. However, camera trajectory errors directly distort human motion, even when camera-space body estimates are accurate.

\begin{figure*}[t]
  \centering
  \includegraphics[width=\textwidth]{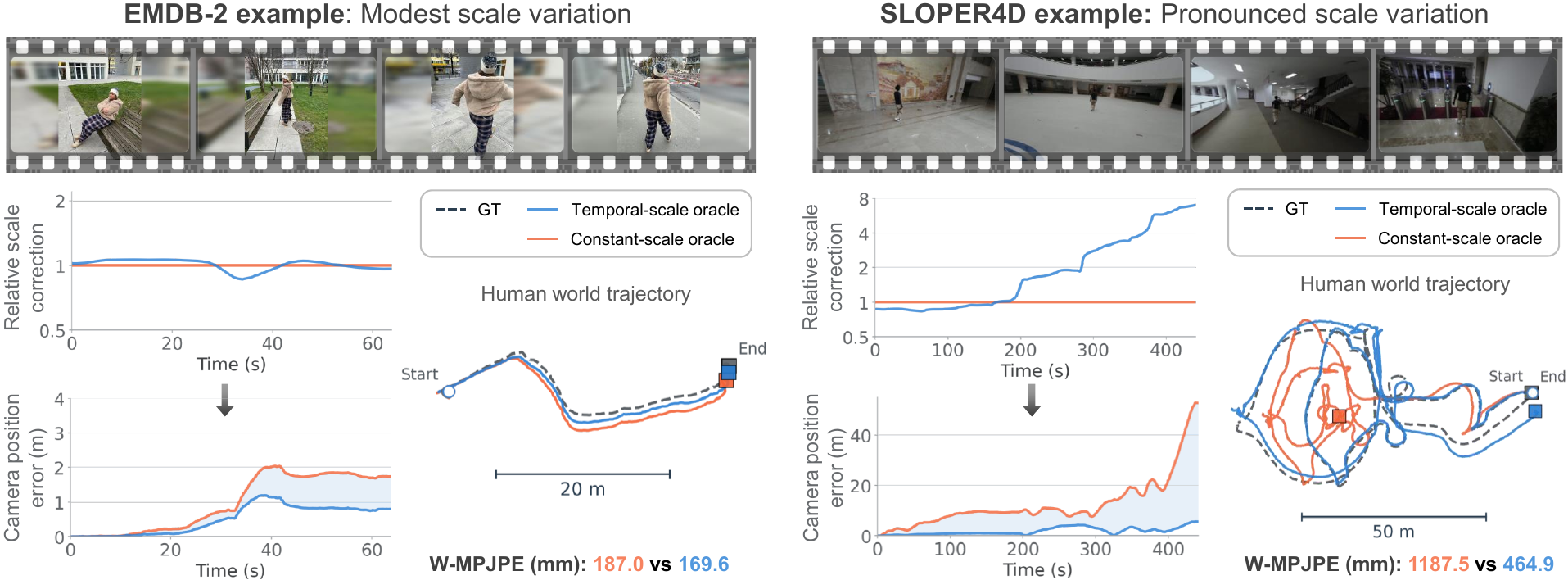}
  \caption{\textbf{Oracle scale comparison on EMDB-2 and SLOPER4D examples.} Constant-scale and temporal-scale oracles are fitted using GT camera trajectories and used to lift identical camera-space body estimates on EMDB-2 (left) and SLOPER4D (right). Temporal correction reduces camera error and improves agreement with GT human trajectories in both examples, with a larger benefit in the SLOPER4D example exhibiting pronounced scale variation. See the supplementary material for oracle construction details and additional oracle results.}
  \label{fig:oracle}
\end{figure*}

Metric scale is a central obstacle to maintaining this spatial consistency. Monocular SLAM~\cite{teed2021droid} recovers camera motion and scene geometry in arbitrary scale rather than metric units. A common calibration strategy uses scene depth, human geometry, or motion cues to estimate one scale for the entire sequence~\cite{wang2024tram, yang2025humans, yin2024whac}. This resolves the overall scale ambiguity, but assumes that the same correction remains valid throughout the camera path. Local scale distortion in practical monocular reconstruction~\cite{strasdat2010scale} can instead make the required correction vary over time. As the camera traverses a scene, residual calibration errors can accumulate and distort the lifted human trajectory. \Cref{fig:motivation} illustrates this limitation: a globally calibrated reconstruction can still lose spatial consistency along its path.

Even a single scale fitted to ground-truth camera motion can leave substantial world-space errors. In \cref{fig:oracle}, we construct a constant-scale oracle and a temporal-scale oracle using ground-truth camera trajectories from EMDB-2 and SLOPER4D, then lift identical camera-space body estimates with each. Temporal correction improves both examples, with a particularly pronounced difference along the extended SLOPER4D camera path. In this example, the constant-scale oracle leaves substantial camera error and human trajectory drift, both of which are greatly reduced by the temporal oracle. With body estimates and camera rotations held fixed, this comparison isolates the effect of the scale correction used for lifting.

To recover this time-varying correction directly from video, we introduce \textbf{\method{}}. Its main component, \textbf{Temporal Scale Calibration}, constructs metric scale observations from predicted body geometry and locally fitted ground planes. The challenge is to preserve meaningful temporal variation despite errors in body depth and scene reconstruction. We therefore estimate the correction jointly over time, preserving informative scale variation while downweighting inconsistent observations and coupling neighboring estimates. This allows calibration to adapt along the camera path while reducing the propagation of observation noise into camera motion. We apply the estimated scales to camera translation increments and integrate them into a calibrated trajectory.

\textbf{Camera-Consistent Motion Refinement} then improves body pose and motion using world-space positions and dynamics computed with the calibrated camera. Using pretrained motion predictions~\cite{wei2026htdrefine} and optical flow~\cite{teed2020raft}, we align body projections with 2D keypoints and surface flow, and match joint velocities and accelerations to the predicted cues. With the camera fixed, regularized updates correct pose and local root motion while restricting global rescaling of the initialized trajectory.

We evaluate on EMDB-2 and SLOPER4D, assessing trajectory accuracy and human motion quality. Ablation studies examine the contributions of temporal scale calibration and motion refinement, while experiments with three HMR backbones assess compatibility with different body estimates. Our contributions are:
\begin{itemize}
  \item We identify and quantify the impact of time-varying camera-scale distortion on world-space HMR, revealing persistent trajectory errors that a single sequence-wide scale cannot resolve.
  \item We introduce \method{}, whose temporal scale calibration robustly recovers time-varying camera-scale corrections from human-ground geometry. Camera-consistent motion refinement improves the lifted body motion while restricting global trajectory rescaling.
  \item We demonstrate substantial gains in world-space accuracy and motion quality on EMDB-2 and SLOPER4D, consistently improving three HMR backbones.
\end{itemize}

\section{Related Work}
\label{sec:related}

\subsection{Human Mesh and Motion Recovery}
Human mesh recovery estimates parametric bodies such as SMPL~\cite{SMPL:2015} and SMPL-X~\cite{pavlakos2019expressive} through fitting~\cite{bogo2016keep}, regression~\cite{kanazawa2018end,goel2023humans}, or their combination~\cite{kolotouros2019learning}. Image-based methods exploit kinematics, visibility, image context, and mesh alignment~\cite{li2021hybrik,kocabas2021pare,li2022cliff,zhang2021pymaf}, while video models add temporal information~\cite{kocabas2020vibe}. World-space recovery additionally requires reasoning about human and camera motion. Existing approaches recover global trajectories from learned motion, temporal, and camera cues~\cite{yuan2022glamr,sun2023trace,shin2024wham}, with GVHMR~\cite{shen2024world} using gravity-view coordinates for world-grounded motion recovery. Other methods compose camera-space bodies with metric camera trajectories~\cite{wang2024tram,wang2025prompthmr}, or jointly recover humans and cameras through optimization and scene reconstruction~\cite{ye2023decoupling,zhao2024synergistic}. Human3R~\cite{chen2026human3r} and JOSH~\cite{ICLR2026_fc657b7f} further extend reconstruction to humans, cameras, and surrounding scenes, while DuoMo~\cite{wang2026duomo} uses diffusion models for camera-space estimation and world-space refinement. Our focus is the metric camera reference used to place body estimates consistently in the world.

\subsection{Metric Calibration for World-Space HMR}
Monocular SLAM, including DROID-SLAM~\cite{teed2021droid} and ORB-SLAM2~\cite{mur2017orb}, recovers camera motion only up to scale and therefore requires a metric reference. Prior SLAM work derives such cues from known object geometry~\cite{sucar2018bayesian}, while world-space human recovery exploits scene and human priors. SLAHMR~\cite{ye2023decoupling} combines relative camera estimates with learned human motion priors to resolve scene-scale ambiguity. TRAM~\cite{wang2024tram} aligns reconstructed background geometry with predicted metric depth, while SynCHMR~\cite{zhao2024synergistic} combines camera-frame human estimates with monocular depth for human-aware metric SLAM. Human-based methods further exploit absolute human depth and motion cues: WHAC~\cite{yin2024whac} recovers metric human and camera trajectories from human-derived spatial cues, while HAC~\cite{yang2025humans} calibrates SLAM using human-scene contact depths. HumanBA~\cite{yang2026humanba} incorporates decoupled human motion as additional constraints in bundle adjustment to improve camera estimation.
 
Most of these approaches seek a global metric scale for the reconstructed trajectory. A recent diagnostic study~\cite{chien2026diagnostic} estimates a sequence-level correction from stationary intervals and reports unstable motion with a segment-wise time-varying alternative. In contrast, we jointly estimate a time-varying scale curve from human-ground observations, downweighting inconsistent measurements and coupling neighboring scale variables to accommodate scale variation without freely following observation noise.

\subsection{Human Motion Refinement}
Motion refinement improves initial estimates using temporal, visual, and physical constraints. SmoothNet~\cite{zeng2022smoothnet} reduces pose jitter through temporal filtering, while motion collections such as AMASS~\cite{mahmood2019amass} support learned priors for fitting~\cite{rempe2021humor,li2024coin}. Scene constraints~\cite{hassan2019resolving}, contact reasoning~\cite{huang2022capturing}, and physics-based methods such as PhysCap~\cite{shimada2020physcap} and PhysHMR~\cite{feng2025physhmr} further encourage plausible human motion and interaction. Most closely related to our refinement, HTD-Refine~\cite{wei2026htdrefine} fits motion to stabilized keypoints and pretrained 3D velocities and accelerations, while optical-flow consistency provides complementary image evidence~\cite{davydov2024using}. These cues constrain temporal motion beyond frame-wise pose accuracy and are complementary to the calibrated camera geometry. We reuse these dynamics cues with surface-flow alignment under a fixed calibrated camera, while body updates allow pose and local root corrections without rescaling the initialized world trajectory.
\section{Method}
\label{sec:method}

\begin{figure*}[t]
  \centering
  \includegraphics[width=\textwidth]{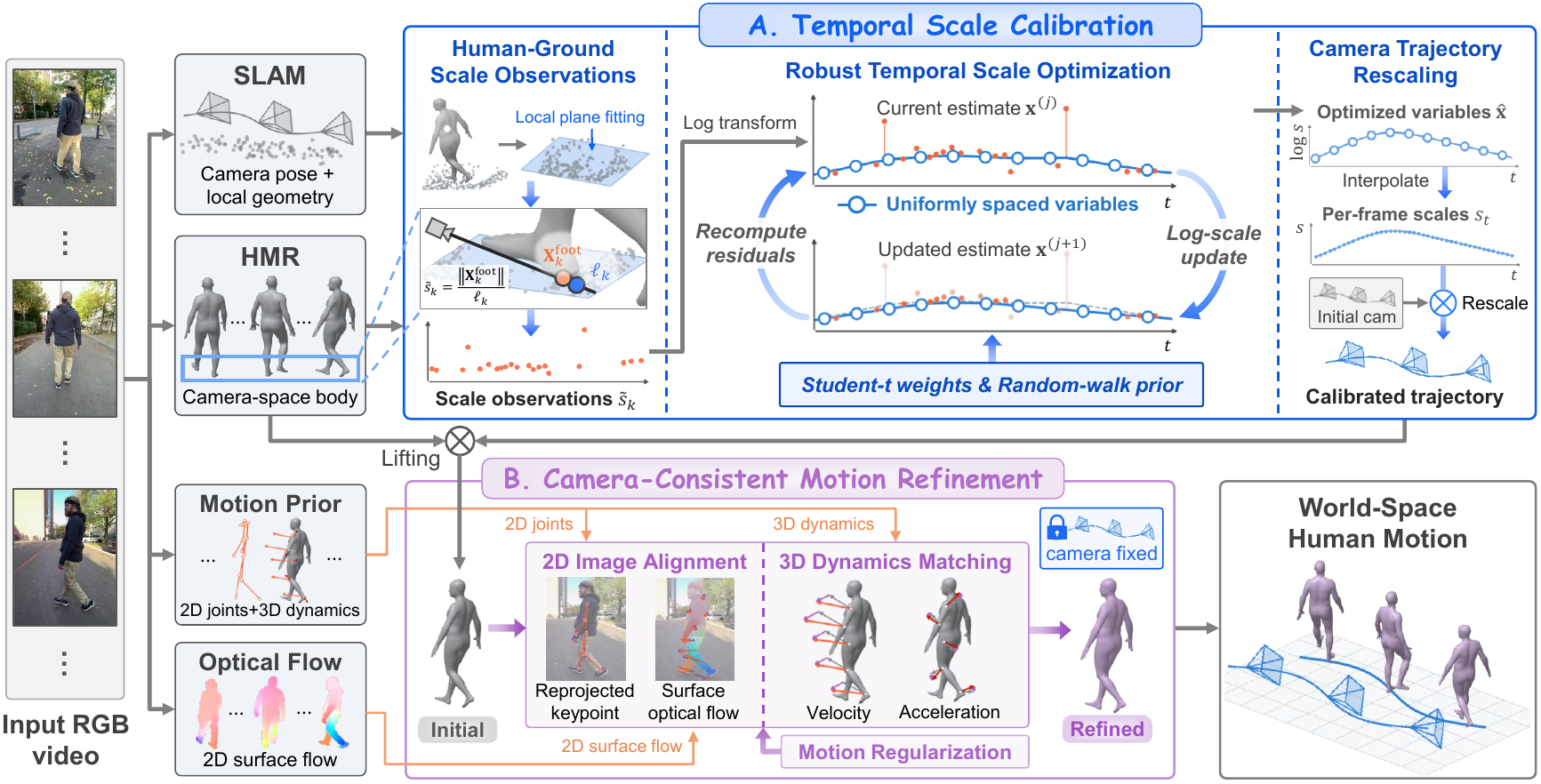}
  \caption{\textbf{Overview of \method{}.} SLAM provides camera poses and scene geometry, while HMR provides camera-coordinate body estimates. \textbf{(A) Temporal Scale Calibration} forms metric scale observations from human-ground geometry at available keyframes and estimates a time-varying correction through robust temporal optimization. The estimated scales calibrate camera translation increments, which are integrated into a metric camera trajectory to lift the body into world coordinates. \textbf{(B) Camera-Consistent Motion Refinement} then improves the lifted motion using 2D image evidence and pretrained 3D dynamics while keeping the calibrated camera fixed.}
  \label{fig:overview}
\end{figure*}

\subsection{Problem Setup and Overview}
\label{sec:setup}

Given a monocular video $\{I_t\}_{t=1}^{T}$ of $T$ frames, our goal is to recover metric human motion in a consistent world coordinate system. An HMR backbone predicts a human mesh in metric camera coordinates, while monocular SLAM provides point clouds $\widetilde{\mathcal P}_k$ at keyframes indexed by $k$ and camera extrinsics $\{(\mathbf R_t,\widetilde{\mathbf t}_t)\}_{t=1}^{T}$, where $\mathbf R_t$ and $\widetilde{\mathbf t}_t$ denote camera-to-world rotation and translation. SLAM reconstructs scene geometry and camera translation at an arbitrary scale, which must be calibrated before combining them with the metric body estimates. Using the calibrated metric camera translation $\mathbf t_t$, a camera-space human body $\mathbf X_t^c$ is lifted to world coordinates as $\mathbf X_t^w=\mathbf R_t\mathbf X_t^c+\mathbf t_t$.
A common approach~\cite{wang2024tram, yang2025humans, zhao2024synergistic, yin2024whac, yang2026humanba} to metric scale calibration is to estimate a single scale $s>0$ for the entire sequence, yielding $\mathbf t_t=\mathbf t_1+s(\widetilde{\mathbf t}_t-\widetilde{\mathbf t}_1)$ for a fixed initial camera position $\mathbf t_1$. To address local scale distortion, we instead estimate a time-varying correction $s_t>0$ and reconstruct the trajectory as $\mathbf t_{t+1}=\mathbf t_t+s_t(\widetilde{\mathbf t}_{t+1}-\widetilde{\mathbf t}_t)$, keeping camera rotations unchanged.

As illustrated in \cref{fig:overview}, \emph{Temporal Scale Calibration} (\cref{sec:camera}) estimates these time-varying corrections from human-ground observations through robust temporal optimization. The calibrated camera trajectory then lifts the body estimates into a world-space initialization. Finally, \emph{Camera-Consistent Motion Refinement} (\cref{sec:human}) refines the body pose and motion using 2D image evidence and 3D dynamics cues while keeping the calibrated camera fixed.

\subsection{Temporal Scale Calibration}
\label{sec:camera}

\paragraph{Human-Ground Scale Observations.}
Inspired by HAC~\cite{yang2025humans}, we use predicted human geometry as a metric reference, with the supporting ground providing a scene correspondence for scale calibration. At each keyframe \(k\) selected by SLAM~\cite{teed2021droid}, we use the lowest projected candidate among predefined foot vertices as a support proxy and locate an image patch around and below it. We collect valid scene points \(\mathcal N_k\subset\widetilde{\mathcal P}_k\) from this patch, excluding human-mask pixels, and fit a local ground plane by least squares. In the keyframe’s SLAM camera coordinates, the plane satisfies \(\mathbf n_k^\top\mathbf q+d_k=0\), where \(\mathbf q\) denotes a point on the plane, \(\mathbf n_k\) is its unit normal, and \(d_k\) is its offset. We retain fits with sufficient, well-spread, near-coplanar points. This local construction adapts to supporting-surface changes along the trajectory.

Let \(\mathbf X_k^{\mathrm{foot}}\) denote the selected foot vertex in metric camera coordinates, with unit ray direction \(\mathbf u_k=\mathbf X_k^{\mathrm{foot}}/\|\mathbf X_k^{\mathrm{foot}}\|\). We compute the ground intersection range and the corresponding scale observation $\widetilde s_k$ as:
\begin{equation}
\ell_k =
-\frac{d_k}{\mathbf n_k^\top\mathbf u_k},
\qquad
\widetilde s_k =
\frac{\|\mathbf X_k^{\mathrm{foot}}\|}{\ell_k}.
\end{equation}
Here, \(\ell_k>0\) is the distance from the camera to the ground-plane intersection in SLAM units, while \(\|\mathbf X_k^{\mathrm{foot}}\|\) is the predicted metric foot distance along the same ray. We then discard observations for which the viewing ray is nearly parallel to the fitted plane. The remaining keyframe observations provide local metric scale cues for nearby camera motion, robustly combined through temporal optimization.

\paragraph{Robust Temporal Scale Optimization.}
Even after geometric filtering, the retained scale observations can still inherit errors from predicted body depth and reconstructed scene geometry. We first compensate for an overall scale bias by forming log-scale observations \(y_k=\log\widetilde s_k-\kappa\), where \(\kappa\) is a fixed log-scale offset. We then estimate the time-varying correction jointly over time to limit the influence of the remaining observation errors.

We represent the correction using \(M\) log-scale variables \(\mathbf x=[x_1,\ldots,x_M]^\top\), uniformly spaced at intervals of \(\Delta=1\,\mathrm{s}\). Their timestamps are fixed, while their values are jointly optimized. This separates the model’s temporal resolution from irregular SLAM keyframe sampling and retains all valid observations at their original timestamps. At each keyframe \(k\), we combine the two neighboring log-scale variables using linear interpolation weights \(\boldsymbol\alpha_k\in\mathbb R^M\), giving the estimated log-scale \(\boldsymbol\alpha_k^\top\mathbf x\).

We use a Student-\(t\) likelihood~\cite{aravkin2014robust}, whose heavy tails reduce the influence of large residuals without a hard rejection threshold. To allow scale to vary over time while discouraging abrupt changes, we couple neighboring log-scale variables with a random-walk prior. We solve the resulting problem using iteratively reweighted least squares. The initial estimate \(\mathbf x^{(0)}\) is obtained by fitting the observations with equal weights and the same prior. At iteration \(j\), we compare the interpolated estimates \(\boldsymbol\alpha_k^\top\mathbf x^{(j)}\) with the observations to compute the residual \(r_k^{(j)}\) and the Student-\(t\) observation weight \(w_k^{(j)}\) as:
\begin{equation}
    r_k^{(j)} = y_k-\boldsymbol\alpha_k^\top\mathbf x^{(j)},
    \quad \;\;
    w_k^{(j)} = \frac{\nu+1}{\nu\sigma^2+\bigl(r_k^{(j)}\bigr)^2},
\end{equation}
where \(\sigma\) and \(\nu\) are the Student-\(t\) scale and degrees of freedom. Larger residuals are downweighted, reducing their influence on the next estimate. We estimate \(\kappa,\sigma,\nu\) on non-test calibration data and hold them fixed during inference.

Holding these weights fixed, we jointly update all log-scale variables through weighted least squares with the random-walk prior:
\begin{equation}
    \begin{aligned}
    \mathbf x^{(j+1)} = \arg\min_{\mathbf x}
    \Bigg[
    &\frac12\sum_k w_k^{(j)}\bigl(y_k-\boldsymbol\alpha_k^\top\mathbf x\bigr)^2\\
    &+ \frac{1}{2q\Delta} \sum_{i=1}^{M-1}(x_{i+1}-x_i)^2
    \Bigg].
    \end{aligned}
\end{equation}
The first term fits the observations using Student-\(t\) weights; the second imposes temporal regularization, with \(q\) controlling the random-walk variance per unit time. We recompute residuals and weights after each update and repeat until numerical convergence, keeping observation values and timestamps fixed. The optimized log-scale variables \(\widehat{\mathbf x}\) are then used for camera trajectory reconstruction.

\paragraph{Camera Trajectory Rescaling.}
After optimization, we linearly interpolate \(\widehat{\mathbf x}\) at each video frame and exponentiate to obtain per-frame scales. We apply these scales to increments of the raw SLAM camera translations \(\widetilde{\mathbf t}_t\) and integrate them into a calibrated trajectory as:
\begin{equation}
    \begin{aligned} 
    s_t = \exp\!\left(\mathbf b_t^\top\widehat{\mathbf x}\right), 
    \quad \;
    \mathbf t_{t+1} = \mathbf t_t + s_t\left(\widetilde{\mathbf t}_{t+1}-\widetilde{\mathbf t}_t\right),
    \end{aligned}
\end{equation}
where \(\mathbf b_t\in \mathbb R^M\) contains the corresponding linear interpolation weights at frame \(t\). Integration starts from the fixed initial position \(\mathbf t_1\), while scale correction leaves the camera rotations \(\mathbf R_t\) unchanged. Applying scale to increments makes the correction independent of SLAM’s arbitrary translation origin. The calibrated camera poses then lift the camera-space body estimates into a world-space initialization for motion refinement.

\subsection{Camera-Consistent Motion Refinement}
\label{sec:human}

Using the calibrated camera, we initialize world-space motion as \(\mathbf X_t^w=\mathbf R_t\mathbf X_t^c+\mathbf t_t\) and refine body parameters \(\boldsymbol\Theta\) comprising pose, global orientation, and root translation, keeping the camera and body shape fixed. Following HTD-Refine~\cite{wei2026htdrefine}, we fit the body to 2D keypoint and 3D dynamics predictions from its pretrained PVA-Net, supplemented by surface optical flow from RAFT~\cite{teed2020raft}. With both networks frozen, refinement combines 2D image alignment and 3D dynamics matching with regularized body updates.

\paragraph{2D Image Alignment.}
2D keypoints align the body with each image, while surface flow connects its motion across frames. Let \(\mathbf X^w_{t,i}\) and \(\mathbf V^w_{t,v}\) denote the current world-space joints and surface vertices determined by \(\boldsymbol\Theta\), and let \(\Pi_t\) denote perspective projection into frame \(t\) through the calibrated camera. We fit PVA keypoints \(\mathbf u^{\mathrm{obs}}_{t,i}\) and RAFT flow \(\mathbf F^{\mathrm{obs}}_{t,v}\), evaluated at initial vertex projections, using:
\begin{equation}
    \begin{aligned}
    \mathbf e^{\mathrm{kp}}_{t,i}
    &=
    \Pi_t(\mathbf X^w_{t,i})-\mathbf u^{\mathrm{obs}}_{t,i},\\
    \mathbf e^{\mathrm{flow}}_{t,v}
    &=
    \Pi_{t+1}(\mathbf V^w_{t+1,v})
    -\Pi_t(\mathbf V^w_{t,v})
    -\mathbf F^{\mathrm{obs}}_{t,v}.
    \end{aligned}
\end{equation}
These residuals align projected joints with keypoints and projected surface motion with optical flow, using the calibrated camera poses to account for camera motion.

\paragraph{3D Dynamics Matching.}
Image agreement alone does not ensure correct 3D motion. We therefore match joint velocities and accelerations to PVA predictions. To separate body dynamics from camera motion, we compute temporal differences in world coordinates and rotate them into the corresponding camera orientations:
\begin{equation}
    \begin{aligned}
    \mathbf e^{\mathrm{vel}}_{t,i}
    &=
    f\mathbf R_t^\top
    (\mathbf X^w_{t+1,i}-\mathbf X^w_{t,i})
    -\mathbf v^{\mathrm{obs}}_{t,i},\\
    \mathbf e^{\mathrm{acc}}_{t,i}
    &=
    f^2\mathbf R_t^\top
    (\mathbf X^w_{t+1,i}-2\mathbf X^w_{t,i}+\mathbf X^w_{t-1,i})
    -\mathbf a^{\mathrm{obs}}_{t,i},
    \end{aligned}
\end{equation}
where \(f\) is the frame rate and \(\mathbf v^{\mathrm{obs}}_{t,i},\mathbf a^{\mathrm{obs}}_{t,i}\) are PVA targets expressed in camera orientation \(t\), with acceleration targets rotated into the corresponding center-frame orientation.

\paragraph{Motion Regularization.}
We additionally regularize body updates to stabilize fitting under uncertain motion evidence:
\begin{equation}
    \mathcal R(\boldsymbol\Theta)
    = \mathcal R_{\mathrm{img}}
    + \mathcal R_{\mathrm{anchor}}
    + \mathcal R_{\mathrm{smooth}}.
\end{equation}
The image-conditioned term \(\mathcal R_{\mathrm{img}}\) assigns tighter rotation-update priors to poorly resolved joints using depth, keypoint confidence, and joint geometry. It also includes standard translation-update priors and temporal regularization of rotation updates. The anchor term \(\mathcal R_{\mathrm{anchor}}\) penalizes deviations from initial joint velocities more weakly along the viewing ray than perpendicular to it, allowing larger depth corrections. The smoothness term \(\mathcal R_{\mathrm{smooth}}\) penalizes third-order world-joint differences to discourage rapid acceleration changes.

Refining root translation can change the overall scale of the trajectory initialized with the calibrated camera. We therefore restrict global rescaling while allowing local root corrections. Let \(\delta\mathbf p\) stack the root translation updates and \(\mathbf g\) stack the initial root positions after subtracting their temporal mean. Updates proportional to \(\mathbf g\) uniformly expand or contract the trajectory, so we impose \(\mathbf g^\top\delta\mathbf p=0\) to remove this scale-changing component.

\begin{table*}[t]
\centering
\caption{\textbf{Comparison on EMDB-2.} Baseline results are quoted from the original papers, except for values marked $^*$, which we compute using official model releases to supplement unreported metrics. $^{\dagger}$ denotes additional joint scene reconstruction with human-scene contact constraints. Bold and underline denote the best and second-best reported values.}
\label{tab:emdb}
\small
\renewcommand{\arraystretch}{1.16}
\setlength{\tabcolsep}{6pt}
\begin{tabular}{lcccccccc}
\toprule
\textbf{Method} & \textbf{Venue} & W-MPJPE & WA-MPJPE & RTE & MPJVE & MPJAE & Jitter & FS\\
\midrule
TRAM~\cite{wang2024tram} & ECCV 2024 & 222.4 & 76.4 & 1.4 & \;\;\underline{0.4}$^*$ & \;\;8.6$^*$ & \;\;29.5$^*$ & \;\;11.6$^*$\\
GVHMR~\cite{shen2024world} & SIGGRAPH Asia 2024 & 276.5 & 111.0 & 2.0 & \;\;\underline{0.4}$^*$ & \;\;\underline{6.7}$^*$ & 16.7 & \textbf{3.5}\\
PromptHMR-Vid~\cite{wang2025prompthmr} & CVPR 2025 & 216.5 & 71.0 & 1.3 & -- & -- & 16.3 & \textbf{3.5}\\
HAC~\cite{yang2025humans} & ICCV 2025 & 197.2 & 71.0 & \underline{1.2} & \;\;\underline{0.4}$^*$ & \;\;8.6$^*$ & \;\;29.5$^*$ & \;\;11.3$^*$\\
GENMO~\cite{li2025genmo} & ICCV 2025 & 202.1 & 74.3 & \underline{1.2} & -- & -- & 17.8 & 8.8\\
JOSH$_3$$^\dagger$~\cite{ICLR2026_fc657b7f} & ICLR 2026 & 174.7 & 68.9 & 1.3 & -- & -- & 9.1 & 12.1\\
HumanBA~\cite{yang2026humanba} & CVPR 2026 & 196.0 & 70.1 & \underline{1.2} & -- & -- & -- & --\\
DuoMo~\cite{wang2026duomo} & CVPR 2026 & \underline{171.6} & \underline{67.6} & \underline{1.2} & -- & -- & 8.7 & \underline{3.7}\\
HTD-Refine~\cite{wei2026htdrefine} & CVPR 2026 & 192.4 & 69.2 & 1.5 & \underline{0.4} & 7.9 & \underline{7.2} & 5.7\\
\rowcolor{oursrow}
\textbf{\method{} (Ours)} & \textbf{This work} & \textbf{154.3} & \textbf{57.3} & \textbf{1.0} & \textbf{0.3} & \textbf{5.4} & \textbf{5.7} & 4.8\\
\bottomrule
\end{tabular}
\end{table*}

\begin{table*}[t]
\centering
\caption{\textbf{Comparison on SLOPER4D.} Baselines marked $^*$ are evaluated by us using official model releases under our protocol; the remaining baseline results are quoted from the original papers. $^{\dagger}$ denotes additional joint scene reconstruction with human-scene contact constraints. Bold and underline denote the best and second-best reported values.}
\label{tab:sloper}
\small
\renewcommand{\arraystretch}{1.16}
\setlength{\tabcolsep}{6pt}
\begin{tabular}{lcccccccc}
\toprule
\textbf{Method} & \textbf{Venue} & W-MPJPE & WA-MPJPE & RTE & MPJVE & MPJAE & Jitter & FS\\
\midrule
TRAM$^{*}$~\cite{wang2024tram} & ECCV 2024 & 999.4 & 174.9 & 5.3 & 1.4 & 32.3 & 101.2 & 62.4\\
GVHMR$^{*}$~\cite{shen2024world} & SIGGRAPH Asia 2024 & 959.1 & 153.7 & 5.2 & 1.1 & 26.2 & 83.2 & 46.0\\
HAC$^{*}$~\cite{yang2025humans} & ICCV 2025 & 1056.4 & 237.7 & 4.0 & 1.3 & 32.1 & 100.7 & 58.3\\
JOSH$_3$$^\dagger$~\cite{ICLR2026_fc657b7f} & ICLR 2026 & \textbf{438.3} & \underline{120.0} & \underline{1.7} & -- & -- & \textbf{7.1} & 28.2\\
HumanBA~\cite{yang2026humanba} & CVPR 2026 & 674.5 & 122.7 & -- & -- & -- & -- & --\\
HTD-Refine$^{*}$~\cite{wei2026htdrefine} & CVPR 2026 & 950.9 & 142.3 & 5.3 & \underline{0.7} & \underline{10.1} & 18.2 & \underline{26.7}\\
\rowcolor{oursrow}
\textbf{\method{} (Ours)} & \textbf{This work} & \underline{446.8} & \textbf{85.0} & \textbf{1.5} & \textbf{0.4} & \textbf{7.1} & \underline{8.7} & \textbf{16.7}\\
\bottomrule
\end{tabular}
\end{table*}

\paragraph{Refinement Objective.}
Under this constraint, we jointly minimize evidence-fitting terms and motion regularization:
{\small
\begin{equation}
    \widehat{\boldsymbol\Theta}
    =
    \arg\min_{\boldsymbol\Theta}
    \left[
    \frac{1}{T}\sum_d\sum_{t,i}
    \omega^d_{t,i}\,
    \rho_d\!\left(\mathbf e^d_{t,i}(\boldsymbol\Theta)\right)
    +
    \mathcal R(\boldsymbol\Theta)
    \right],
\end{equation}}
where \(d\in\{\mathrm{kp},\mathrm{flow},\mathrm{vel},\mathrm{acc}\}\) specifies the residual type, \(t\) ranges over valid frames, and \(i\) indexes joints or surface vertices. Here, \(\omega^d_{t,i}\) combines relative term weights, confidence, and validity masks, while \(\rho_d\) is a Cauchy penalty for robust fitting. We use the L-BFGS optimizer to jointly update the body parameters over the full sequence, obtaining refined world-space motion \(\widehat{\mathbf X}_t^w\).

\section{Experiments}
\label{sec:experiments}

\subsection{Experimental Setup}

\paragraph{Datasets.}
We evaluate on two moving-camera benchmarks for world-space human recovery. EMDB-2~\cite{kaufmann2023emdb} is a widely used benchmark, on which we evaluate all 25 sequences. SLOPER4D~\cite{dai2023sloper4d} tests spatial consistency over extended camera paths, where local calibration errors can accumulate into large trajectory deviations. We evaluate six SLOPER4D sequences under a shared protocol.

\vspace{-0.7em}
\paragraph{Evaluation metrics.}
We assess world-space joint accuracy using Mean Per-Joint Position Error with initial-segment alignment (W-MPJPE, mm) and full-segment alignment (WA-MPJPE, mm). Root Translation Error (RTE, \%) measures aligned root-trajectory error normalized by GT path length, while Absolute Trajectory Error (ATE, m) measures camera-position error after rigid alignment. Mean Per-Joint Velocity Error (MPJVE, m/s) and Acceleration Error (MPJAE, m/s$^2$) measure world-space motion accuracy. Jitter (10m/s$^3$) and Foot Sliding (FS, mm) assess temporal smoothness and foot stability, respectively.

\vspace{-0.7em}
\paragraph{Implementation details.}
\label{sec:implementation}
Unless otherwise stated, we use GVHMR~\cite{shen2024world} as the body backbone and DROID-SLAM~\cite{teed2021droid} for camera reconstruction. Temporal scale calibration uses \(\Delta=1\,\mathrm{s}\) and \(q=1.6\times10^{-4}\), with \(\kappa,\sigma,\nu\) calibrated on BEDLAM2.0 training sequences~\cite{tesch2026bedlam2}. All pretrained networks remain frozen. For motion refinement, the relative coefficients for keypoint, flow, velocity, and acceleration fitting are \(1,0.25,1,1\), respectively, combined with observation-specific confidence and validity weights. We optimize body parameters using L-BFGS for at most 3,000 iterations, with shared settings across both datasets. Detailed evaluation protocols, implementation settings, and exact parameter values, including \(\kappa,\sigma,\nu\), are provided in the supplementary material.

\begin{table}[t]
\centering
\caption{\textbf{Ablation of Temporal Scale Calibration on SLOPER4D.} Arrows denote replacing the left-hand component with the right-hand alternative, while ``w/o” denotes removal. The same human motion refinement is applied to all variants.}
\label{tab:camera}
\small
\renewcommand{\arraystretch}{1.16}
\setlength{\tabcolsep}{2.5pt}
\begin{tabular*}{\linewidth}{@{\hspace{5pt}\extracolsep{\fill}}lccc@{\hspace{5pt}}}
\toprule
\textbf{Variant} & ATE & W-MPJPE & RTE\\
\midrule
\rowcolor{tablegroup}
\multicolumn{4}{@{\hspace{5pt}}l}{\strut\textbf{\textit{Robust Temporal Scale Optimization}}}\\
\;\;Temporal $\rightarrow$ Constant scale & 7.52 & 708.12 & 2.93\\
\;\;Optimization $\rightarrow$ Gaussian smoothing & 5.86 & 503.77 & 1.87\\
\;\;Student-$t$ $\rightarrow$ Gaussian likelihood & 5.47 & 481.39 & 1.73\\
\;\;w/o random-walk prior & 5.30 & 522.37 & 1.89\\
\midrule
\rowcolor{tablegroup}
\multicolumn{4}{@{\hspace{5pt}}l}{\strut \textbf{\textit{Human-Ground Scale Observations}}}\\
\;\;Plane-ray $\rightarrow$ Below-foot cue & 5.22 & 473.09 & 1.78\\
\midrule
\textbf{Full calibration} & \textbf{4.36} & \textbf{446.83} & \textbf{1.47}\\
\bottomrule
\end{tabular*}
\end{table}

\begin{table}[t]
\centering
\caption{\textbf{Ablation of Camera-Consistent Motion Refinement on SLOPER4D.} ``Camera only'' excludes human motion refinement, while ``w/o'' denotes component removal. All variants share the same calibrated camera and initialization.}
\label{tab:human}
\small
\renewcommand{\arraystretch}{1.16}
\setlength{\tabcolsep}{2.5pt}
\begin{tabular*}{\linewidth}{@{\hspace{5pt}\extracolsep{\fill}}lccc@{\hspace{5pt}}}
\toprule
\textbf{Variant} & W-MPJPE & MPJVE & MPJAE\\
\midrule
Camera only & 473.76 & 0.95 & 26.01\\
\midrule
w/o 2D image alignment & 465.22 & 0.44 & \textbf{7.10}\\
w/o 3D dynamics matching & 455.04 & 0.48 & 7.41\\
w/o motion regularization & 494.97 & 0.68 & 14.90\\
\midrule
\textbf{Full refinement} & \textbf{446.83} & \textbf{0.43} & 7.12\\
\bottomrule
\end{tabular*}
\end{table}

\subsection{Comparison with the State of the Art}

\paragraph{Results on EMDB-2.}
The EMDB-2 comparison in \cref{tab:emdb} shows a clear advance in world-space joint accuracy. W-MPJPE reaches 154.3mm, improving over the strongest baseline, DuoMo, by 10.1\%, while WA-MPJPE improves by 15.2\%. The advantage therefore persists after full-segment alignment, indicating better reconstruction of the motion within each segment. Accurate joint recovery is accompanied by a low root-trajectory error and favorable dynamics and jitter results, showing that the gains extend from spatial placement to motion quality.

\vspace{-0.8em}
\paragraph{Results on SLOPER4D.}
The benefit becomes particularly pronounced along SLOPER4D's extended camera paths. Compared with the GVHMR baseline in \cref{tab:sloper}, \method{} reduces W-MPJPE from 959.1 to 446.8mm, a 53.4\% improvement from the same initial body estimates. Improvements across the remaining metrics show that more accurate trajectories also yield better motion quality. Across all reported results, \method{} achieves the lowest WA-MPJPE, RTE, and FS. Its W-MPJPE is also within 1.9\% of the published JOSH$_3$ result, demonstrating competitive world-space accuracy through temporal camera calibration and body refinement without joint scene optimization.

\begin{table}[t]
\centering
\caption{\textbf{Body-backbone compatibility on SLOPER4D.} Each body backbone uses the same initial TRAM camera trajectory.}
\label{tab:backbones}
\small
\renewcommand{\arraystretch}{1.16}
\setlength{\tabcolsep}{2.5pt}
\begin{tabular*}{\linewidth}{@{\hspace{5pt}\extracolsep{\fill}}lcccc@{\hspace{5pt}}}
\toprule
\textbf{Backbone} & W-MPJPE & RTE & MPJVE & MPJAE\\
\midrule
TRAM~\cite{wang2024tram} & 999.45 & 5.26 & 1.35 & 32.29\\
\quad + \textbf{\method{}} & \textbf{489.00} & \textbf{1.50} & \textbf{0.48} & \textbf{7.53}\\
\midrule
Human3R~\cite{chen2026human3r} & 1345.14 & 5.58 & 1.73 & 39.14\\
\quad + \textbf{\method{}} & \textbf{870.27} & \textbf{1.86} & \textbf{0.69} & \textbf{8.77}\\
\midrule
GVHMR~\cite{shen2024world} & 959.08 & 5.22 & 1.09 & 26.18\\
\quad + \textbf{\method{}} & \textbf{446.83} & \textbf{1.47} & \textbf{0.43} & \textbf{7.12}\\
\bottomrule
\end{tabular*}
\end{table}

\subsection{Ablation Studies}
\label{sec:ablation}
We conduct controlled ablations on SLOPER4D, where extended camera paths make accumulated calibration errors easier to observe in the recovered motion.

\vspace{-0.5em}
\paragraph{Temporal Scale Calibration.}
\Cref{tab:camera} examines calibration accuracy through camera ATE and its downstream effects through W-MPJPE and RTE, with trajectory integration and human refinement fixed. The largest loss of accuracy occurs when the correction is forced to remain constant: W-MPJPE rises from 446.83 to 708.12mm despite retaining the same Student-\(t\) likelihood. How temporal variation is estimated also matters. Interpolating the same log-scale observations to every frame and applying Gaussian smoothing with a 1-s bandwidth still yields 503.77mm W-MPJPE. Within the optimization, replacing Student-\(t\) with a Gaussian likelihood or removing the random-walk prior also increases error, supporting complementary roles for robust weighting and temporal coupling. The below-foot control further supports the proposed observation construction. The full formulation achieves the lowest errors across all three metrics. Additional controls and bandwidth sensitivity analyses are provided in the supplementary material.

\begin{figure*}[t]
  \centering
  \includegraphics[width=\textwidth]{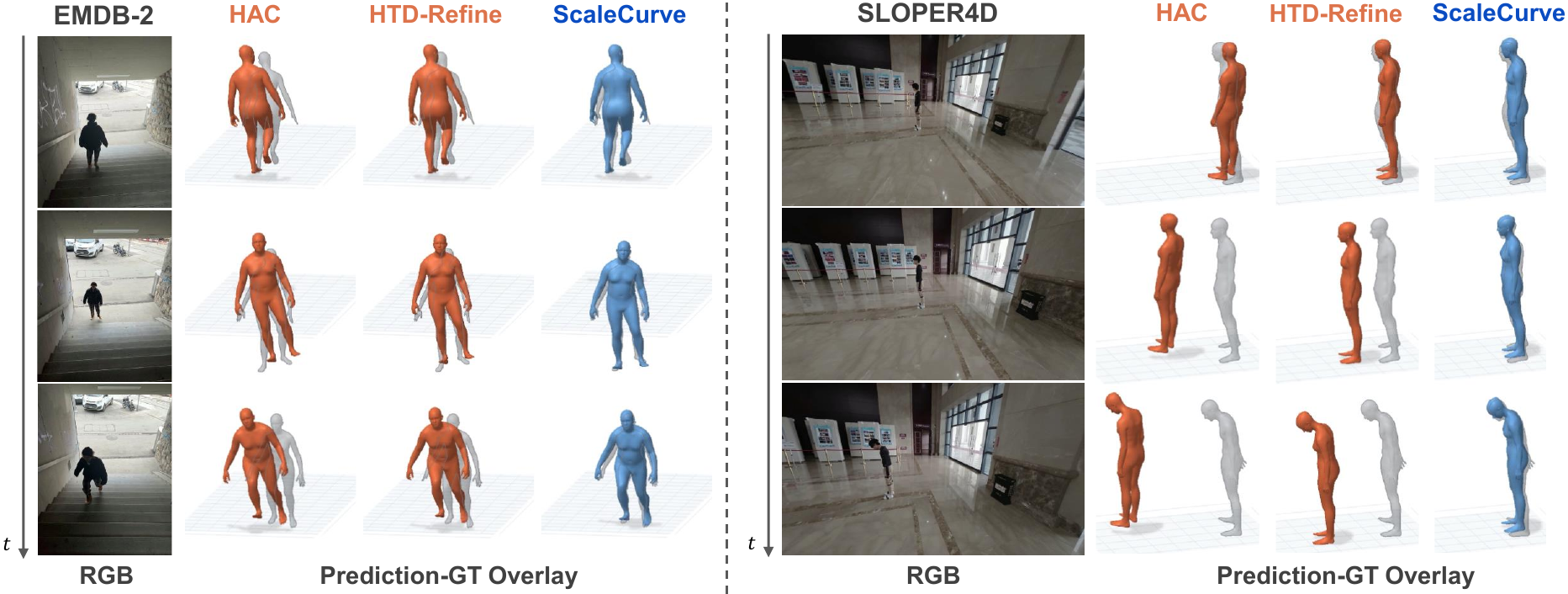}
  \caption{\textbf{Qualitative comparison of human motion.} Reconstructed human meshes from HAC~\cite{yang2025humans}, HTD-Refine~\cite{wei2026htdrefine}, and \method{} are overlaid with GT (gray) in world coordinates on EMDB-2 (left) and SLOPER4D (right). Time progresses downward.}
  \label{fig:qual_mesh}
\end{figure*}

\begin{figure}[t]
  \centering
  \includegraphics[width=\linewidth]{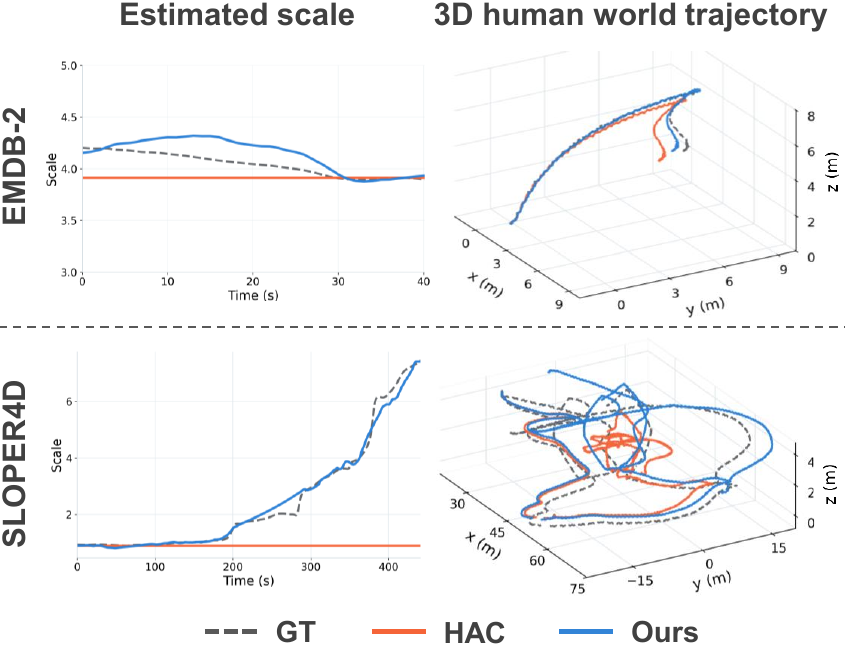}
  \caption{\textbf{Qualitative comparison of scale and trajectories.} Estimated scales (left) and world-space human trajectories (right) from HAC~\cite{yang2025humans} and \method{} on EMDB-2 (top) and SLOPER4D (bottom). Gray dashed curves indicate the GT-assisted scale oracle and GT trajectories, respectively.}
  \label{fig:qual_trajectory}
  \vspace{-1.0em}
\end{figure}

\vspace{-0.5em}
\paragraph{Camera-Consistent Motion Refinement.}
Once the camera is calibrated, refinement delivers substantial gains in motion dynamics. In \cref{tab:human}, the full stage reduces velocity and acceleration errors by 54.7\% and 72.6\% relative to camera-only initialization, while also improving world-space joint accuracy. The component controls reveal complementary roles: 2D image alignment primarily improves body placement, while 3D dynamics matching improves temporal motion accuracy. Our motion regularization is particularly important: removing its penalties and the root-scale constraint more than doubles acceleration error to 14.90m/s$^2$ and raises W-MPJPE to 494.97mm, exceeding the camera-only initialization. Regularized updates therefore help turn motion evidence into improvements in both dynamics and world-space accuracy.

\vspace{-0.5em}
\paragraph{Compatibility with different body backbones.}
The benefits remain consistent across HMR backbones with different initial accuracies. In \cref{tab:backbones}, we apply \method{} to TRAM~\cite{wang2024tram}, Human3R~\cite{chen2026human3r}, and GVHMR~\cite{shen2024world}, all using the same initial TRAM camera trajectory. All four metrics improve, with W-MPJPE reductions of 35.3\%--53.4\%, demonstrating compatibility with diverse body predictions while keeping each backbone frozen.

Additional ablations on EMDB-2, together with further experimental results and detailed analyses, are provided in the supplementary material.

\subsection{Qualitative Results}
\label{sec:qualitative}
The qualitative comparisons in \cref{fig:qual_mesh} reveal the effects of inconsistent world-space placement. In the EMDB-2 stair example, HAC and HTD-Refine depart from GT in height as the subject ascends, whereas \method{} follows the elevation change more closely. In the SLOPER4D example, the subject remains in place while the camera moves, yet the baselines drift and exhibit apparent foot sliding. Our reconstruction keeps the body and feet substantially more stable. This illustrates how residual camera-scale errors can appear as human displacement, highlighting the importance of calibrating the world reference before refining body motion.

The scale and trajectory plots in \cref{fig:qual_trajectory} connect the estimated correction to its downstream effect. In the EMDB-2 example, our correction captures the temporal oracle’s later decrease, yielding a trajectory closer to GT in horizontal extent and elevation. In the SLOPER4D example, the required correction increases substantially along the camera path. Our estimate follows this temporal pattern, while HAC retains a constant scale. The resulting human trajectory follows the GT route and its wider excursions more faithfully. Additional qualitative results, including videos, are provided in the supplementary material.

\section{Conclusion}
\label{sec:conclusion}

We presented \method{}, a world-space HMR framework that treats camera scale calibration as temporal estimation. Robust estimation from human-ground geometry corrects scale variation, while camera-consistent refinement improves body motion with restricted global rescaling. Experiments on EMDB-2 and SLOPER4D show substantial gains in trajectory accuracy and motion quality across three HMR backbones. It requires usable ground geometry and does not correct rotation errors. These results highlight the need for calibration beyond global scale ambiguity and offer a practical route toward more reliable human-scene interaction analysis without backbone retraining.

{
    \small
    \bibliographystyle{ieeenat_fullname}
    \bibliography{main}
}

\clearpage
\appendix

\begin{center}
    {\Large\bfseries Supplementary Material}
\end{center}
\vspace{0.5em}

\noindent 
This supplementary material provides further details and analyses of \method{}. We first describe the implementation and evaluation protocols, then examine temporal scale calibration and motion refinement through GT-assisted scale diagnostics, component ablations, and parameter sensitivity studies. We also evaluate compatibility with different body backbones and camera reconstruction methods, report computational costs, and present additional qualitative results and limitations.

\section{Implementation Details}
\label{sec:implementation}
This section details the implementation of temporal scale calibration and camera-consistent motion refinement. Each experiment runs on a single NVIDIA RTX 6000 Ada GPU. \Cref{tab:hyperparameters} summarizes the calibrated observation parameters and fixed settings used in both stages.

\begin{table*}[t]
\centering
\caption{\textbf{Calibration parameters and fixed settings.} The observation parameters are estimated on BEDLAM2.0 training data and held fixed during inference; no additional network is trained. Other settings are shared across EMDB-2 and SLOPER4D unless noted. Image residual scales are specified at $1280\times960$ resolution.}
\label{tab:hyperparameters}
\small
\setlength{\tabcolsep}{6pt}
\begin{tabular*}{\linewidth}{@{\hspace{10pt}\extracolsep{\fill}}lccc@{\hspace{5pt}}}
\toprule
\rowcolor{tablegroup}
\multicolumn{4}{@{\hspace{5pt}}l}{\strut\textbf{\textit{Calibrated observation parameters}}}\\
\textbf{Observation configuration} & \textbf{Bias offset $\kappa$} & \textbf{Noise scale $\sigma$} & \textbf{Degrees of freedom $\nu$}\\
\midrule
SLOPER4D configuration & $-0.005409$ & $0.049418$ & $1.318450$\\
EMDB-2 configuration & $-0.033496$ & $0.050579$ & $1.070191$\\
\midrule
\rowcolor{tablegroup}
\multicolumn{4}{@{\hspace{5pt}}l}{\strut\textbf{\textit{Fixed camera-stage settings}}}\\
\textbf{Setting} & \multicolumn{3}{l}{\textbf{Value}}\\
\midrule
Frame rate; temporal interval $\Delta$ & \multicolumn{3}{l}{30 fps; 1 s}\\
Random-walk variance rate $q$ & \multicolumn{3}{l}{$1.6\times10^{-4}$}\\
Patch radius; minimum plane support & \multicolumn{3}{l}{24 disparity pixels; 30 points}\\
Coplanarity; spread thresholds & \multicolumn{3}{l}{0.25; 0.30}\\
IRLS maximum iterations; tolerance & \multicolumn{3}{l}{500; $10^{-8}$}\\
Numerical anchor $\epsilon$ & \multicolumn{3}{l}{$10^{-6}$}\\
\midrule
\rowcolor{tablegroup}
\multicolumn{4}{@{\hspace{5pt}}l}{\strut\textbf{\textit{Fixed human-stage settings}}}\\
\textbf{Setting} & \multicolumn{3}{l}{\textbf{Value}}\\
\midrule
Keypoint / flow / velocity / acceleration weights & \multicolumn{3}{l}{1 / 0.25 / 1 / 1}\\
Keypoint / flow Cauchy scales & \multicolumn{3}{l}{5 / 1 pixels, scaled with image resolution}\\
Velocity / acceleration Cauchy scales & \multicolumn{3}{l}{0.098212 m/s / 1.269494 m/s$^2$}\\
Pose-prior multiplier; smoothness scale $\sigma_a$ & \multicolumn{3}{l}{2; 3}\\
Translation / rotation-update standard deviations & \multicolumn{3}{l}{0.05 m / 0.02 rad}\\
GVHMR ray / perpendicular anchor deviations & \multicolumn{3}{l}{0.557329 / 0.190055 m/s}\\
L-BFGS maximum iterations; history size & \multicolumn{3}{l}{3,000; 20}\\
L-BFGS gradient / change tolerances & \multicolumn{3}{l}{$10^{-9}$ / $10^{-12}$}\\
\bottomrule
\end{tabular*}
\end{table*}

\subsection{Frozen Inputs and Coordinate Conventions}
We process $T$ video frames at $f=30$ fps; $t$ and $k$ index frames and SLAM keyframes. Unless otherwise stated, DROID-SLAM~\cite{teed2021droid} supplies camera poses and disparities through TRAM~\cite{wang2024tram}, while GVHMR~\cite{shen2024world} supplies the default camera-space body estimates. PVA-Net from HTD-Refine~\cite{wei2026htdrefine} provides keypoints and 3D dynamics, and RAFT~\cite{teed2020raft} provides optical flow. All networks and body shape remain fixed. Scale observations are constructed from the reconstructed scene geometry and the predicted foot geometry of the corresponding HMR backbone.

Camera-to-world rotations are $\mathbf R_t$, and raw and metric camera translations are $\widetilde{\mathbf t}_t$ and $\mathbf t_t$, respectively. Positive scales $s_t$ rescale translation increments while preserving rotations. After integration, one rigid transformation sets the first camera to the origin with identity orientation. We use dataset-provided camera intrinsics; test GT bodies and poses are used only for evaluation and oracle diagnostics.

\subsection{Human-Ground Scale Observations}
At each keyframe, we select the lowest projected foot vertex from SMPL indices 3216, 3387, 6617, and 6787. We collect background points around and below it with a 24-disparity-pixel patch radius, excluding invalid disparities and person-mask pixels. Least-squares plane fitting requires at least 30 points, a smallest-to-largest singular-value ratio below 0.25, and a middle-to-largest ratio of at least 0.30, ensuring approximately coplanar, well-spread support.

Let $\mathbf n_k,d_k$ be the plane's unit normal and offset in SLAM camera coordinates, and $\mathbf X_k^{\mathrm{foot}}$ the metric camera-space foot vertex. Its unit ray $\mathbf u_k=\mathbf X_k^{\mathrm{foot}}/\|\mathbf X_k^{\mathrm{foot}}\|$ intersects the plane at range $\ell_k=-d_k/(\mathbf n_k^\top\mathbf u_k)$ in SLAM units. We retain intersections with $\ell_k>10^{-3}$ and $|\mathbf n_k^\top\mathbf u_k|>10^{-3}$, yielding the raw scale observation $\widetilde s_k=\|\mathbf X_k^{\mathrm{foot}}\|/\ell_k$.

\subsection{Robust Temporal Scale Optimization}
The bias-corrected log-scale observation is $y_k=\log\widetilde s_k-\kappa$, with fixed offset $\kappa$. We optimize $M$ log-scale variables $\mathbf x=[x_1,\ldots,x_M]^\top$ at uniform intervals $\Delta=1$ s. Interpolation weights $\boldsymbol\alpha_k\in\mathbb R^M$ connect keyframe $k$ to its two neighboring variables, giving the estimate $\boldsymbol\alpha_k^\top\mathbf x$. The parameter $q$ is the random-walk variance per unit time.

With residual $r_k=y_k-\boldsymbol\alpha_k^\top\mathbf x$, the scale objective underlying the iteratively reweighted least-squares (IRLS) updates is:
\begin{equation}
\begin{split}
E_{\mathrm{scale}}(\mathbf x)={}&\frac{\nu+1}{2}\sum_k\log\left(1+\frac{r_k^2}{\nu\sigma^2}\right)\\
&+\frac{1}{2q\Delta}\sum_{i=1}^{M-1}(x_{i+1}-x_i)^2+\frac{\epsilon}{2}x_1^2,
\end{split}
\end{equation}
where $\sigma,\nu$ are the Student-$t$ scale and degrees of freedom, and $\epsilon=10^{-6}$ weakly anchors the first variable for numerical stability. The parameters $\kappa,\sigma,\nu$ are calibrated on BEDLAM2.0 training sequences~\cite{tesch2026bedlam2} and fixed during inference.

Starting from equal observation weights, IRLS alternates reweighting and tridiagonal solves using the stopping criteria in \cref{tab:hyperparameters}. Observation values and timestamps remain fixed. The optimized variables $\widehat{\mathbf x}$ yield per-frame scales $s_t=\exp(\mathbf b_t^\top\widehat{\mathbf x})$, where $\mathbf b_t\in\mathbb R^M$ contains interpolation weights at frame $t$.

\subsection{Camera-Consistent Motion Refinement}
We refine body parameters $\boldsymbol\Theta$ (pose, global orientation, and root translation) after lifting with the calibrated camera, keeping camera poses and shape fixed. SMPL-X bodies are fitted through a fixed-shape SMPL proxy; updates are transferred back to the native body representation for evaluation.

\paragraph{Image and dynamics evidence.}
For keypoint, flow, velocity, and acceleration residuals, indexed by $d\in\{\mathrm{kp},\mathrm{flow},\mathrm{vel},\mathrm{acc}\}$, we use $\rho_d(\mathbf e)=\log(1+\|\mathbf e\|^2/\eta_d^2)$, where $\mathbf e$ is the residual vector and $\eta_d$ its Cauchy scale. Weights $\omega^d_{t,i}$ combine the coefficients in \cref{tab:hyperparameters} with confidence and validity masks for joint or vertex $i$. For image height $H$ and width $W$ in pixels, image-residual scales are multiplied by $\sqrt{HW/(1280\cdot960)}$. World-joint differences are compared in PVA camera orientations, using the center frame for acceleration.

RAFT flow is converted to full-resolution pixels and queried at initial vertex projections. Surface anchors, visibility, and forward--backward consistency masks remain fixed. Projected surface displacements use the calibrated camera pose at each endpoint.

\paragraph{Regularization.}
The image-conditioned term \(\mathcal R_{\mathrm{img}}\) tightens priors using joint depth, focal length, keypoint confidence, and rest-pose joint-to-keypoint distances, ignoring confidences below 0.3. Rotation-prior standard deviations are clipped to 0.01–0.2 radians, then doubled for keypoint-connected joints. Translation-update and second-difference rotation-update priors are included.

The anchor term \(\mathcal R_{\mathrm{anchor}}\) allows larger velocity deviations from initialization along the viewing ray than perpendicular to it. We prevent uniform trajectory rescaling through \(\mathbf g^\top\delta\mathbf p=0\), where \(\delta\mathbf p\) stacks root translation updates and \(\mathbf g\) stacks centered initial root positions.

For motion smoothness, let $\mathbf X_t^w$ stack the current world-space joint positions and let $D^3\mathbf X_t^w=\mathbf X_{t+3}^w-3\mathbf X_{t+2}^w+3\mathbf X_{t+1}^w-\mathbf X_t^w$ denote their third temporal difference. The implemented term is:
\begin{equation}
\mathcal R_{\mathrm{smooth}}=\frac{3}{10\sigma_a^2T}\sum_{t=1}^{T-3}\|f^2D^3\mathbf X_t^w\|^2,
\end{equation}
where $\sigma_a=3$ controls the penalty on consecutive acceleration changes $f^2D^3\mathbf X_t^w$.

We use full-sequence L-BFGS with strong-Wolfe line search and the shared settings in \cref{tab:hyperparameters}. The accompanying code provides full prior construction, input formats, and inference commands.
    
\section{Evaluation Protocols}
\label{sec:evaluation}
We detail the alignment, preprocessing, and aggregation used for evaluation.

\paragraph{Datasets and valid frames.}
We evaluate all 25 EMDB-2 sequences~\cite{kaufmann2023emdb} and six SLOPER4D sequences~\cite{dai2023sloper4d}. Local EMDB-2 comparisons use official valid-frame masks, apply corrections to joints and vertices, and average errors over evaluated samples. SLOPER4D predictions are interpolated from 30 fps to 20 Hz annotation timestamps without translation smoothing. Joint metrics use contiguous valid runs of at least 200 frames, weighted by run length within sequences and averaged equally across sequences. Controlled comparisons share valid frames. Oracle diagnostics use unsmoothed lifted bodies; both oracle metrics and camera ATE use equal sequence averaging.

\paragraph{World-space accuracy.}
W-MPJPE and WA-MPJPE measure mean per-joint position error in millimeters. Each 100-frame segment is aligned using its first two frames for W-MPJPE or the full segment for WA-MPJPE; both alignments include scale. Root Translation Error (RTE, \%) normalizes aligned root-trajectory error by GT path length. Camera Absolute Trajectory Error (ATE, m) measures root-mean-square camera-position error after full-sequence rigid alignment without scale fitting.

\paragraph{Motion quality.}
MPJVE (m/s) and MPJAE (m/s$^2$) measure world-joint velocity and acceleration error after rigid alignment. Temporal differences use physical timestamps and exclude invalid intervals. Foot Sliding (FS, mm) measures displacement of four foot vertices whose GT speed is below 0.3m/s. Local Jitter averages $\|f^3D^3\mathbf X^w_{t,i}\|/10$ over valid frames $t$ and joints $i$, where $\mathbf X^w_{t,i}$ is a world-space joint, $D^3$ the third temporal difference, and $f$ the evaluation frame rate. Its reporting unit is $10\,\mathrm{m/s^3}$; published values retain their original conventions.

Main-paper comparisons use published results where available and mark locally evaluated metrics. Supplementary studies evaluate official model releases under controlled settings to isolate our components’ effects; baseline values may therefore differ from main-paper Table~1.
\section{GT-Assisted Scale Diagnostics}
\label{sec:oracles}
Figure~2 in the main paper compares constant-scale and temporal-scale oracles on one example from each benchmark. Here we describe their construction and extend the comparison to all 25 EMDB-2 and six SLOPER4D sequences in \cref{tab:oracle}. Both oracles use GT camera trajectories while lifting identical camera-space body estimates with unchanged camera rotations. They are diagnostic references, not inputs to \method{}.

\paragraph{Oracle construction.}
The constant-scale oracle uses the scale of a whole-sequence similarity fit between SLAM and GT camera positions. The temporal-scale oracle obtains local similarity-fit scales from overlapping windows centered at one-second intervals. We interpolate the resulting log-scales to video frames, exponentiate, and apply them to camera translation increments. Each local fit requires sufficient valid positions and nondegenerate motion; when a sequence has insufficient valid local fits, its oracle reverts to the constant scale. Window durations follow the recorded effective bandwidth of the reference camera-filtering protocol and may differ by sequence. These fitting windows are distinct from the one-second variable interval used by \method{}; the temporal oracle does not fit an independent free scale at every frame.

\paragraph{Aggregate comparison.}
\Cref{tab:oracle} reports camera ATE after full-path rigid alignment and W-MPJPE on the common diagnostic support, averaged equally across sequences. This ATE differs from the start-camera-aligned error curves in the main figure. Temporal correction reduces both aggregate errors on both datasets, with a larger difference on SLOPER4D. Because the body estimates and camera rotations remain fixed, the comparison isolates the correction used for lifting. Neither oracle is a global lower bound on human reconstruction error: rotation, displacement-direction, and body errors remain uncorrected.

\begin{table}[t]
\centering
\caption{\textbf{Oracle scale comparison across both benchmarks.} Camera ATE and W-MPJPE are sequence averages over all 25 EMDB-2 and six SLOPER4D sequences. Both oracles lift identical bodies on the same diagnostic support.}
\label{tab:oracle}
\small
\renewcommand{\arraystretch}{1.16}
\setlength{\tabcolsep}{3pt}
\begin{tabular*}{\linewidth}{@{\hspace{5pt}\extracolsep{\fill}}lcc@{\hspace{5pt}}}
\toprule
\textbf{Oracle} & ATE (m) & W-MPJPE (mm) \\\midrule
\rowcolor{tablegroup}
\multicolumn{3}{@{\hspace{5pt}}l}{\textit{\textbf{EMDB-2 (25 sequences)}}} \\
Constant-scale oracle & 0.33 & 177.19 \\
Temporal-scale oracle & 0.24 & 168.71 \\
\midrule
\rowcolor{tablegroup}
\multicolumn{3}{@{\hspace{5pt}}l}{\textit{\textbf{SLOPER4D (6 sequences)}}} \\
Constant-scale oracle & 7.18 & 745.09 \\
Temporal-scale oracle & 1.89 & 360.32 \\
\bottomrule
\end{tabular*}
\end{table}
\section{Additional Analyses}
\label{sec:ablation_extra}

\begin{table*}[t]
\centering
\caption{\textbf{Complementarity of the two stages.} TSC: Temporal Scale Calibration; CMR: Camera-Consistent Motion Refinement. Crosses disable the corresponding stage; the first row lifts GVHMR with the TRAM camera. All variants share the same initial body and camera estimates. Both datasets use the main evaluation protocols.}
\label{tab:stage_factorial}
\small
\renewcommand{\arraystretch}{1.16}
\setlength{\tabcolsep}{3pt}
\begin{tabular*}{\linewidth}{@{\hspace{5pt}\extracolsep{\fill}}lcccccccccc@{\hspace{5pt}}}
\toprule
 & & & \multicolumn{4}{c}{\textbf{EMDB-2}} & \multicolumn{4}{c}{\textbf{SLOPER4D}} \\
\cmidrule(lr){4-7}\cmidrule(lr){8-11}
\textbf{Variant} & TSC & CMR & W-MPJPE & RTE & MPJVE & MPJAE & W-MPJPE & RTE & MPJVE & MPJAE \\
\midrule
GVHMR + TRAM camera & $\times$ & $\times$ & 202.16 & 1.32 & 0.34 & 6.43 & 959.08 & 5.22 & 1.09 & 26.18\\
Calibration only & $\checkmark$ & $\times$ & 163.00 & \textbf{0.99} & 0.34 & 6.43 & 473.76 & 1.49 & 0.95 & 26.01\\
Refinement only & $\times$ & $\checkmark$ & 196.54 & 1.32 & 0.32 & \textbf{5.45} & 933.86 & 5.21 & 0.62 & 7.20\\
\midrule
\textbf{Full \method{}} & $\checkmark$ & $\checkmark$ & \textbf{154.35} & \textbf{0.99} & \textbf{0.31} & \textbf{5.45} & \textbf{446.83} & \textbf{1.47} & \textbf{0.43} & \textbf{7.12}\\
\bottomrule
\end{tabular*}
\end{table*}

\subsection{Complementarity of Calibration and Refinement}
\label{sec:stage_combination}
\Cref{tab:stage_factorial} examines the complementary roles of Temporal Scale Calibration (TSC) and Camera-Consistent Motion Refinement (CMR) on both datasets. All four variants use the same camera-space GVHMR estimates and sequence-wide calibrated TRAM camera as initialization. We enable each stage independently, so refinement operates with the initial camera when TSC is disabled and with the temporally calibrated camera in the full system. Body inputs and evaluation settings remain shared across the comparison.

Camera calibration provides the main improvement in trajectory accuracy, while refinement contributes strongly to motion dynamics. On SLOPER4D, calibration alone reduces W-MPJPE from 959.08 to 473.76mm before any body refinement. Refinement alone reduces acceleration error from 26.18 to 7.20m/s$^2$, yet leaves W-MPJPE at 933.86mm. This contrast shows that fitting motion evidence cannot fully compensate for an inaccurate camera reference. Applying refinement after calibration further improves W-MPJPE to 446.83mm while achieving an acceleration error of 7.12m/s$^2$. On EMDB-2, the combined system likewise achieves the lowest W-MPJPE at 154.35mm. Together, these results support the two-stage design: calibration establishes a more accurate world-space reference, and refinement improves the lifted motion within it.

\subsection{Temporal Scale Calibration}
\label{sec:camera_extra}
Camera-stage studies keep the camera-space body inputs fixed. Component ablations include the same downstream human refinement as the main paper, while sensitivity studies and geometric diagnostics evaluate camera calibration directly using ATE.

\paragraph{Component ablation on EMDB-2.}
\Cref{tab:camera_emdb} extends main-paper Table~3 to all 25 EMDB-2 sequences, with the same downstream human refinement across variants. Temporal calibration reduces W-MPJPE from 168.97 to 154.35mm compared with a constant scale, extending its benefit beyond SLOPER4D. Replacing optimization with Gaussian smoothing at a one-second bandwidth produces substantially larger errors, showing that smoothing observations alone is insufficient. The Gaussian-likelihood and no-random-walk controls further support robust observation weighting and temporal coupling, respectively. Plane–ray observations also improve world-space joint accuracy over below-foot cues. These results reinforce the main-paper findings on a second benchmark.

\begin{table}[t]
\centering
\caption{\textbf{Temporal Scale Calibration ablation on EMDB-2.} All variants include the same human motion refinement and use the main evaluation protocol. Arrows indicate replacement; w/o denotes removal.}
\label{tab:camera_emdb}
\small
\renewcommand{\arraystretch}{1.16}
\setlength{\tabcolsep}{2pt}
\begin{tabular*}{\linewidth}{@{\hspace{5pt}\extracolsep{\fill}}lccc@{\hspace{5pt}}}
\toprule
\textbf{Variant} & ATE & W-MPJPE & RTE\\
\midrule
\rowcolor{tablegroup}
\multicolumn{4}{@{\hspace{5pt}}l}{\strut\textbf{\textit{Robust Temporal Scale Optimization}}}\\
\;\;Temporal $\rightarrow$ Constant scale & 0.70 & 168.97 & 1.13\\
\;\;Optimization $\rightarrow$ Gaussian smoothing & 4.45 & 319.38 & 7.71\\
\;\;Student-$t$ $\rightarrow$ Gaussian likelihood & 0.80 & 173.96 & 1.36\\
\;\;w/o random-walk prior & 6.35 & 379.54 & 11.04\\
\midrule
\rowcolor{tablegroup}
\multicolumn{4}{@{\hspace{5pt}}l}{\strut\textbf{\textit{Human-Ground Scale Observations}}}\\
\;\;Plane-ray $\rightarrow$ Below-foot cue & 0.63 & 171.69 & 1.12 \\
\midrule
\textbf{Full calibration} & \textbf{0.58} & \textbf{154.35} & \textbf{0.99}\\
\bottomrule
\end{tabular*}
\end{table}

\begin{figure*}[t]
\centering
\includegraphics[width=.9\textwidth]{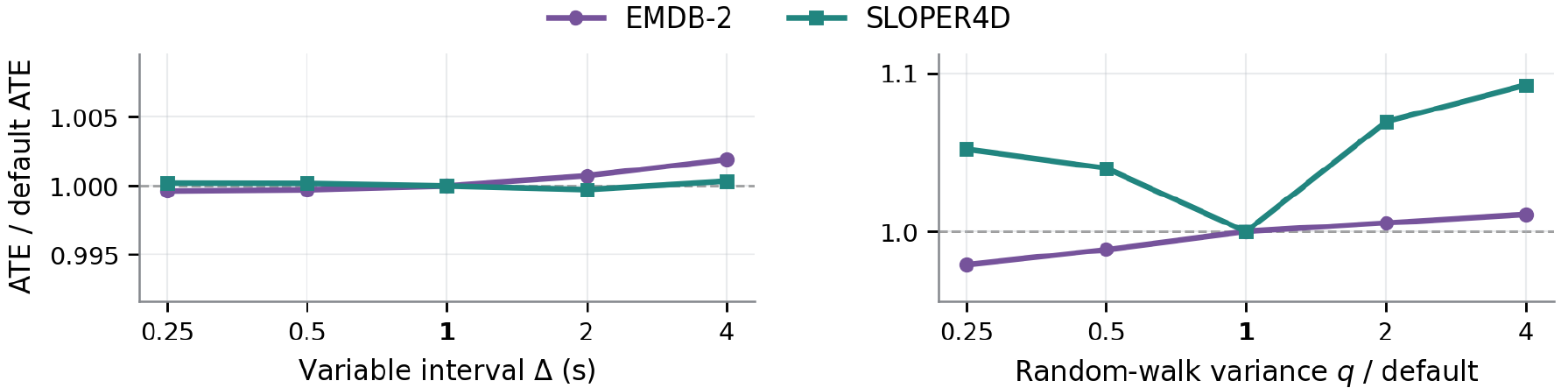}
\caption{\textbf{Sensitivity of temporal scale calibration.} Variable interval $\Delta$ and random-walk variance rate $q$ are varied independently, with all other settings fixed. Camera ATE is normalized by the default within each dataset. The $q$ axis scales the default $1.6\times10^{-4}$; bold ticks mark the main-paper settings.}
\label{fig:sensitivity}
\end{figure*}

\paragraph{Temporal parameter sensitivity.}
\Cref{fig:sensitivity} examines two settings governing temporal scale estimation: the log-scale variable interval $\Delta$ and random-walk variance rate $q$, which determine temporal resolution and coupling strength, respectively. \Cref{tab:hyperparameters} lists the remaining settings, including calibrated observation parameters and motion-refinement weights. We vary each setting independently, holding all others fixed, including $\kappa,\sigma,\nu$ at their calibration values. Camera accuracy changes little over intervals from 0.25 to 4 seconds, whereas changing $q$ has a larger effect on SLOPER4D. These curves assess sensitivity around the shared settings used in the main paper.

\begin{figure*}[t]
\centering
\includegraphics[width=.95\textwidth]{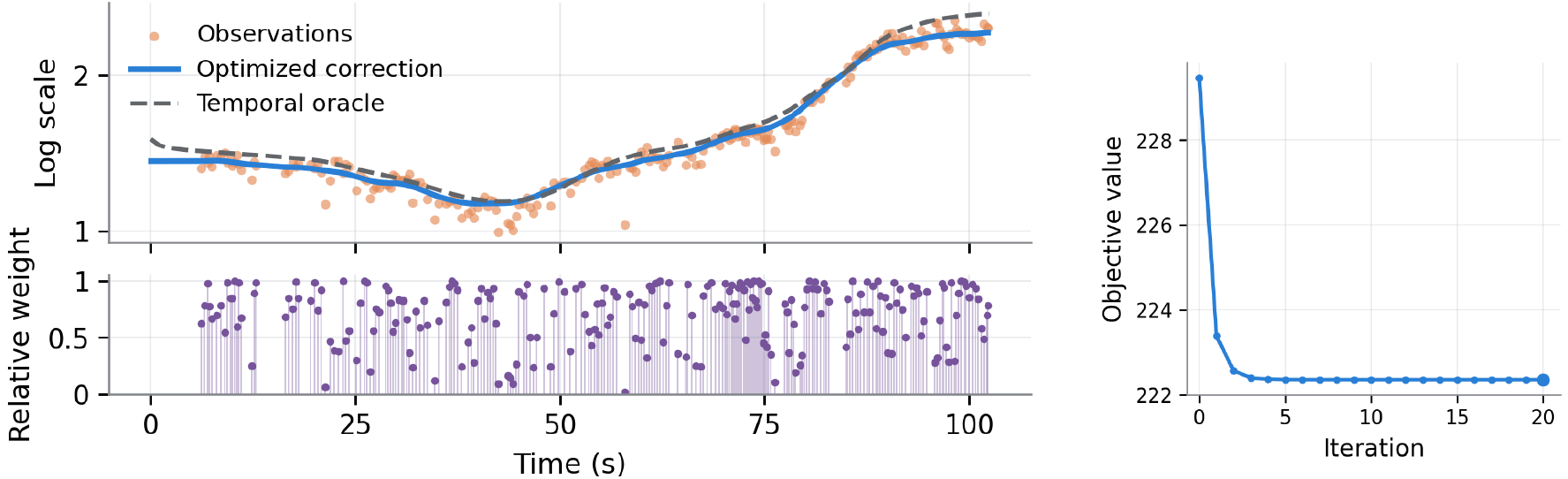}
\caption{\textbf{Scale observations, robust weights, and convergence.} Left: log-scale observations and the optimized correction on SLOPER4D Running (top), with Student-$t$ weights normalized by their zero-residual value (bottom). The dashed curve is the GT-assisted temporal oracle. Right: the optimization objective over IRLS iterations; iteration zero is the equal-weight initialization. }
\label{fig:observation_weights}
\end{figure*}

\paragraph{Observation influence in a real sequence.}
\Cref{fig:observation_weights} illustrates temporal scale optimization on the SLOPER4D sequence. Observations far from the fitted curve receive less influence, while temporal coupling preserves the changing scale trend. The GT-assisted temporal oracle is shown only for reference and is not used by the optimizer. The solver converges after 20 IRLS updates from the equal-weight initialization, with observation values and timestamps fixed throughout.

\begin{figure*}[t]
\centering
\includegraphics[width=.9\textwidth]{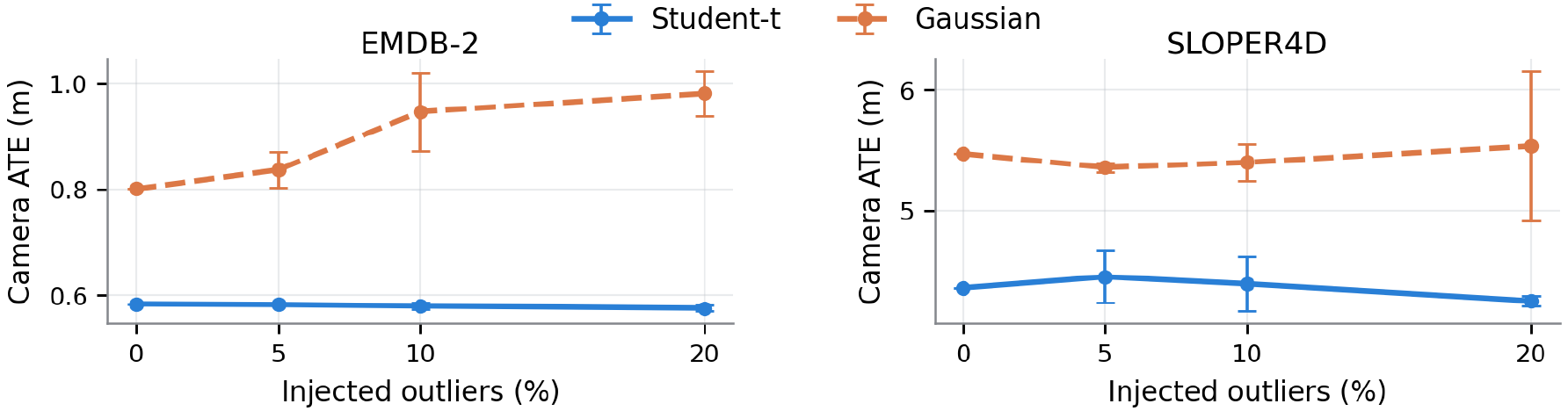}
\caption{\textbf{Robustness to contaminated scale observations.} Camera ATE under identical added outliers, with all other calibration settings fixed. Error bars show standard deviation across three seeds after averaging sequences. Zero contamination uses the original observations.}
\label{fig:robustness}
\end{figure*}

\paragraph{Robustness to additional outliers.}
To assess robustness to unreliable scale observations, we perturb 5\%, 10\%, or 20\% of the retained log-scale observations by a randomly signed \(\log 2\). We compare Student-\(t\) and Gaussian likelihoods under identical perturbations, averaging results over all sequences and three random seeds in \cref{fig:robustness}. At 20\% contamination, Student-\(t\) yields lower ATE on both datasets: 0.575 versus 0.981m on EMDB-2 and 4.250 versus 5.533m on SLOPER4D. These results show that robust weighting limits the influence of corrupted observations while retaining useful scale evidence. The test addresses observation outliers rather than prolonged loss of ground support. Both likelihoods use the same 500-iteration limit, and runs reaching this limit remain included.

\begin{table}[t]
\centering
\caption{\textbf{Comparison of scale application rules.} Camera ATE (m), averaged across sequences, for scaling translation increments versus absolute camera positions. Both use identical estimated scales and camera rotations.}
\label{tab:integration}
\small
\renewcommand{\arraystretch}{1.16}
\setlength{\tabcolsep}{4pt}
\begin{tabular*}{\linewidth}{@{\hspace{5pt}\extracolsep{\fill}}lcc@{\hspace{5pt}}}
\toprule
\textbf{Rescaling rule} & EMDB-2 ATE & SLOPER4D ATE \\\midrule
Translation increments & 0.583 & 4.358 \\
Absolute positions & 0.819 & 38.814 \\
\bottomrule
\end{tabular*}
\end{table}

\paragraph{Applying time-varying scale corrections.}
Our method reconstructs the calibrated camera trajectory by scaling translation increments and integrating them over time. In \cref{tab:integration}, we compare this rule with directly scaling absolute camera positions, keeping the estimated scales and camera rotations fixed. Direct position scaling produces larger camera ATE, particularly on SLOPER4D. Scaling increments preserves each displacement’s direction and adjusts its magnitude, allowing the corrected motion to accumulate into a calibrated trajectory. It also avoids dependence on SLAM’s arbitrary coordinate origin, supporting its use for applying time-varying scale corrections.

\begin{table*}[t]
\centering
\caption{\textbf{Ablation of motion regularization.} Each removal is relative to full refinement. The calibrated camera, initialization, motion evidence, optimization budget, and root-scale constraint are shared. }
\label{tab:regularization_terms}
\small
\renewcommand{\arraystretch}{1.16}
\setlength{\tabcolsep}{6pt}
\begin{tabular*}{\linewidth}{@{\hspace{5pt}\extracolsep{\fill}}lcccccc@{\hspace{5pt}}}
\toprule
& \multicolumn{3}{c}{\textbf{EMDB-2}} & \multicolumn{3}{c}{\textbf{SLOPER4D}}\\
\cmidrule(lr){2-4}\cmidrule(lr){5-7}
\textbf{Variant} & W-MPJPE & MPJVE & MPJAE & W-MPJPE & MPJVE & MPJAE\\
\midrule
w/o image-conditioned priors ($\mathcal R_{\mathrm{img}}$) & 165.91 & 0.317 & 5.58 & 483.41 & \textbf{0.422} & \textbf{6.96}\\
w/o velocity anchoring ($\mathcal R_{\mathrm{anchor}}$) & 155.76 & 0.314 & 5.46 & 450.53 & 0.432 & 7.11\\
w/o motion smoothness ($\mathcal R_{\mathrm{smooth}}$) & \textbf{152.44} & 0.316 & 5.81 & 452.57 & 0.727 & 18.47\\
\midrule
\textbf{Full regularization} & 154.35 & \textbf{0.311} & \textbf{5.45} & \textbf{446.83} & 0.430 & 7.12\\
\bottomrule
\end{tabular*}
\end{table*}

\subsection{Camera-Consistent Motion Refinement}
\label{sec:human_extra}
We retain the calibrated camera, original body initialization, shape, and optimization budget while testing the grouped components and individual regularization terms. The full setting uses pose-prior multiplier two, smoothness scale three, and flow weight 0.25 on both datasets.

\paragraph{Grouped ablation on EMDB-2.}
\Cref{tab:human_emdb} extends main-paper Table~4 to EMDB-2. The calibrated camera is identical across rows. Image alignment provides the clearest position improvement: removing it increases W-MPJPE from 154.35 to 178.18mm. Motion regularization, including the root-scale constraint, also improves both position and dynamics. The isolated contribution of 3D dynamics is smaller on this benchmark, showing that the balance between image and dynamics evidence depends on the sequence.

\paragraph{Individual regularization terms.}
\Cref{tab:regularization_terms} isolates the three terms of $\mathcal R=\mathcal R_{\mathrm{img}}+\mathcal R_{\mathrm{anchor}}+\mathcal R_{\mathrm{smooth}}$ on both datasets. Each control removes one complete term while retaining all 2D and 3D fitting terms and the root-scale constraint. Removing $\mathcal R_{\mathrm{img}}$ also removes the basic parameter-update priors grouped with it.

Image-conditioned priors protect world-space joint accuracy: removing them increases W-MPJPE by 11.56mm on EMDB-2 and 36.58mm on SLOPER4D. Motion smoothness has the clearest effect on dynamics, with its removal raising SLOPER4D MPJAE from 7.12 to 18.47m/s$^2$. Velocity anchoring provides smaller position gains on both datasets. The terms balance position and dynamics: omitting $\mathcal R_{\mathrm{smooth}}$ slightly improves EMDB-2 W-MPJPE but worsens dynamics, while omitting $\mathcal R_{\mathrm{img}}$ modestly improves SLOPER4D dynamics at a higher cost in position accuracy.

\begin{table}[t]
\centering
\caption{\textbf{Camera-Consistent Motion Refinement ablation on EMDB-2.} All 25 EMDB-2 sequences use the main comparison preprocessing. Camera only excludes refinement; each removal is relative to the full stage.}
\label{tab:human_emdb}
\small
\renewcommand{\arraystretch}{1.16}
\setlength{\tabcolsep}{2.5pt}
\begin{tabular*}{\linewidth}{@{\hspace{5pt}\extracolsep{\fill}}lccc@{\hspace{5pt}}}
\toprule
\textbf{Variant} & W-MPJPE & MPJVE & MPJAE\\\midrule
Camera only & 163.00 & 0.34 & 6.43\\
\midrule
w/o 2D image alignment & 178.18 & 0.33 & 5.56\\
w/o 3D dynamics matching & 156.22 & \textbf{0.31} & 5.49\\
w/o motion regularization & 161.39 & 0.33 & 6.25\\
\midrule
\textbf{Full refinement} & \textbf{154.35} & \textbf{0.31} & \textbf{5.45}\\
\bottomrule
\end{tabular*}
\end{table}
\section{Body and SLAM Compatibility}
\label{sec:compatibility}

\begin{table*}[t]
\centering
\caption{\textbf{Compatibility with different camera reconstruction methods.} Each base combines the named camera reconstruction method and its reconstructed geometry with GVHMR; $+$\method{} applies both calibration and human refinement. Each pair uses identical sequence and frame support. W-MPJPE: mm; RTE: \%; MPJVE/MPJAE: m/s and m/s$^2$.}
\label{tab:alternate_slam}
\small
\renewcommand{\arraystretch}{1.16}
\setlength{\tabcolsep}{4pt}
\begin{tabular*}{\linewidth}{@{\hspace{5pt}\extracolsep{\fill}}lcccccccc@{\hspace{5pt}}}
\toprule
 & \multicolumn{4}{c}{\textbf{EMDB-2}} & \multicolumn{4}{c}{\textbf{SLOPER4D}} \\
\cmidrule(lr){2-5}\cmidrule(lr){6-9}
\textbf{Methods} & W-MPJPE & RTE & MPJVE & MPJAE & W-MPJPE & RTE & MPJVE & MPJAE \\
\midrule
ViPE~\cite{huang2025vipe} & 239.12 & 2.41 & 0.36 & 6.44 & 534.39 & 2.64 & 1.07 & 27.50 \\
\quad + \textbf{\method{}} & \textbf{152.44} & \textbf{1.13} & \textbf{0.31} & \textbf{5.44} & \textbf{448.77} & \textbf{1.18} & \textbf{0.46} & \textbf{7.28} \\
\midrule
MegaSaM~\cite{li2025megasam} & 281.41 & 2.95 & 0.41 & 6.61 & 804.32 & 5.13 & 1.30 & 29.03 \\
\quad + \textbf{\method{}} & \textbf{169.86} & \textbf{1.51} & \textbf{0.35} & \textbf{5.69} & \textbf{488.09} & \textbf{2.48} & \textbf{0.65} & \textbf{7.40} \\
\midrule
DROID-SLAM~\cite{teed2021droid} & 202.16 & 1.32 & 0.34 & 6.43 & 959.08 & 5.22 & 1.09 & 26.18 \\
\quad + \textbf{\method{}} & \textbf{154.35} & \textbf{0.99} & \textbf{0.31} & \textbf{5.45} & \textbf{446.83} & \textbf{1.47} & \textbf{0.43} & \textbf{7.12} \\
\bottomrule
\end{tabular*}
\end{table*}

\begin{table}[t]
\centering
\caption{\textbf{Body-backbone compatibility on EMDB-2.} Each backbone uses the same initial TRAM camera. The second row applies the complete \method{} pipeline. Both rows share inputs and benchmark support.}
\label{tab:backbones_emdb}
\small
\renewcommand{\arraystretch}{1.16}
\setlength{\tabcolsep}{2.5pt}
\begin{tabular*}{\linewidth}{@{\hspace{5pt}\extracolsep{\fill}}lcccc@{\hspace{5pt}}}
\toprule
\textbf{Backbone} & W-MPJPE & RTE & MPJVE & MPJAE\\
\midrule
TRAM~\cite{wang2024tram} & 257.16 & 1.32 & 0.40 & 8.62\\
\quad + \textbf{\method{}} & \textbf{197.29} & \textbf{1.13} & \textbf{0.34} & \textbf{6.70}\\
\midrule
Human3R~\cite{chen2026human3r} & 293.27 & 1.45 & 0.60 & 17.25\\
\quad + \textbf{\method{}} & \textbf{250.06} & \textbf{1.09} & \textbf{0.38} & \textbf{8.07}\\
\midrule
GVHMR~\cite{shen2024world} & 202.16 & 1.32 & 0.34 & 6.43\\
\quad + \textbf{\method{}} & \textbf{154.35} & \textbf{0.99} & \textbf{0.31} & \textbf{5.45}\\
\bottomrule
\end{tabular*}
\end{table}

\subsection{Compatibility with Different Camera Reconstruction Methods}
We evaluate whether \method{} generalizes to different camera reconstruction models using ViPE~\cite{huang2025vipe}, MegaSaM~\cite{li2025megasam}, and DROID-SLAM~\cite{teed2021droid}, with GVHMR fixed as the body backbone. This comparison tests compatibility beyond the default camera reconstruction. Each base row lifts the body estimates with the corresponding camera trajectory, while +\method{} applies temporal scale calibration followed by camera-consistent motion refinement. All pairs use the same sequences and matched frame support, retaining the default calibration and refinement settings.

As shown in \cref{tab:alternate_slam}, \method{} improves all four metrics across all reconstruction methods on both datasets. For MegaSaM, W-MPJPE decreases from 281.41 to 169.86mm on EMDB-2 and from 804.32 to 488.09mm on SLOPER4D. These gains also hold for ViPE, showing that camera reconstructions informed by learned metric-depth priors still benefit from our framework. The improvements across all three reconstruction methods demonstrate compatibility beyond the default DROID-SLAM setting.

\subsection{Compatibility with Different HMR Backbones}
\Cref{tab:backbones_emdb} extends main-paper Table~5 to EMDB-2. For each backbone, we combine camera-space predictions from TRAM~\cite{wang2024tram}, Human3R~\cite{chen2026human3r}, or GVHMR~\cite{shen2024world} with the same initial TRAM camera trajectory and evaluate the improvement from adding \method{}. All pairs use the main EMDB-2 evaluation protocol. These comparisons assess compatibility with different body estimates rather than the native published systems in main-paper Table~1.

Adding \method{} improves all four metrics across all three backbones, reducing W-MPJPE by 23.3\%, 14.7\%, and 23.7\%, respectively, alongside improvements in root trajectories and motion dynamics. These consistent gains demonstrate compatibility with diverse body estimates of varying initial accuracy without backbone retraining. Refinement retains shared settings except for backbone-specific directional anchor uncertainties.
\section{Computational Cost}
\label{sec:cost}
We profile \method{} on SLOPER4D Running (3,071 frames) using a single NVIDIA RTX 6000 Ada GPU. \Cref{tab:runtime} separates preprocessing from Temporal Scale Calibration (TSC) and Camera-Consistent Motion Refinement (CMR). Preprocessing includes inference with frozen models such as SLAM, HMR, PVA-Net, and RAFT. It accounts for 98.84\% of the measured runtime, while TSC and CMR optimization require only 0.12\% and 1.04\%, respectively. Notably, TSC includes observation construction, temporal optimization, and trajectory rescaling without introducing an additional neural network. Temporal calibration therefore contributes little computational overhead to the complete pipeline. Full-system throughput is computed from the sum of separately measured stage times.

\begin{table}[t]
\centering
\caption{\textbf{Computational cost of ScaleCurve.} TSC and CMR denote Temporal Scale Calibration and Camera-Consistent Motion Refinement, respectively. FPS denotes stage-specific throughput, while the final row reports full-pipeline throughput. Runtime (\%) indicates each stage's fraction of the total runtime.}
\label{tab:runtime}
\small
\renewcommand{\arraystretch}{1.16}
\setlength{\tabcolsep}{6pt}
\begin{tabular*}{\linewidth}{@{\hspace{5pt}\extracolsep{\fill}}lcc@{\hspace{5pt}}}
\toprule
\textbf{Stage} & \textbf{FPS} & \textbf{Runtime (\%)}\\
\midrule
Preprocessing & 1.26 & 98.84\\
\textbf{TSC} & \textbf{1018.47} & \textbf{0.12}\\
CMR & 119.91 & 1.04\\
\midrule
Full ScaleCurve & 1.25 & 100.00\\
\bottomrule
\end{tabular*}
\end{table}
\begin{figure*}[t]
\centering
\includegraphics[width=.95\textwidth,height=.88\textheight,keepaspectratio]{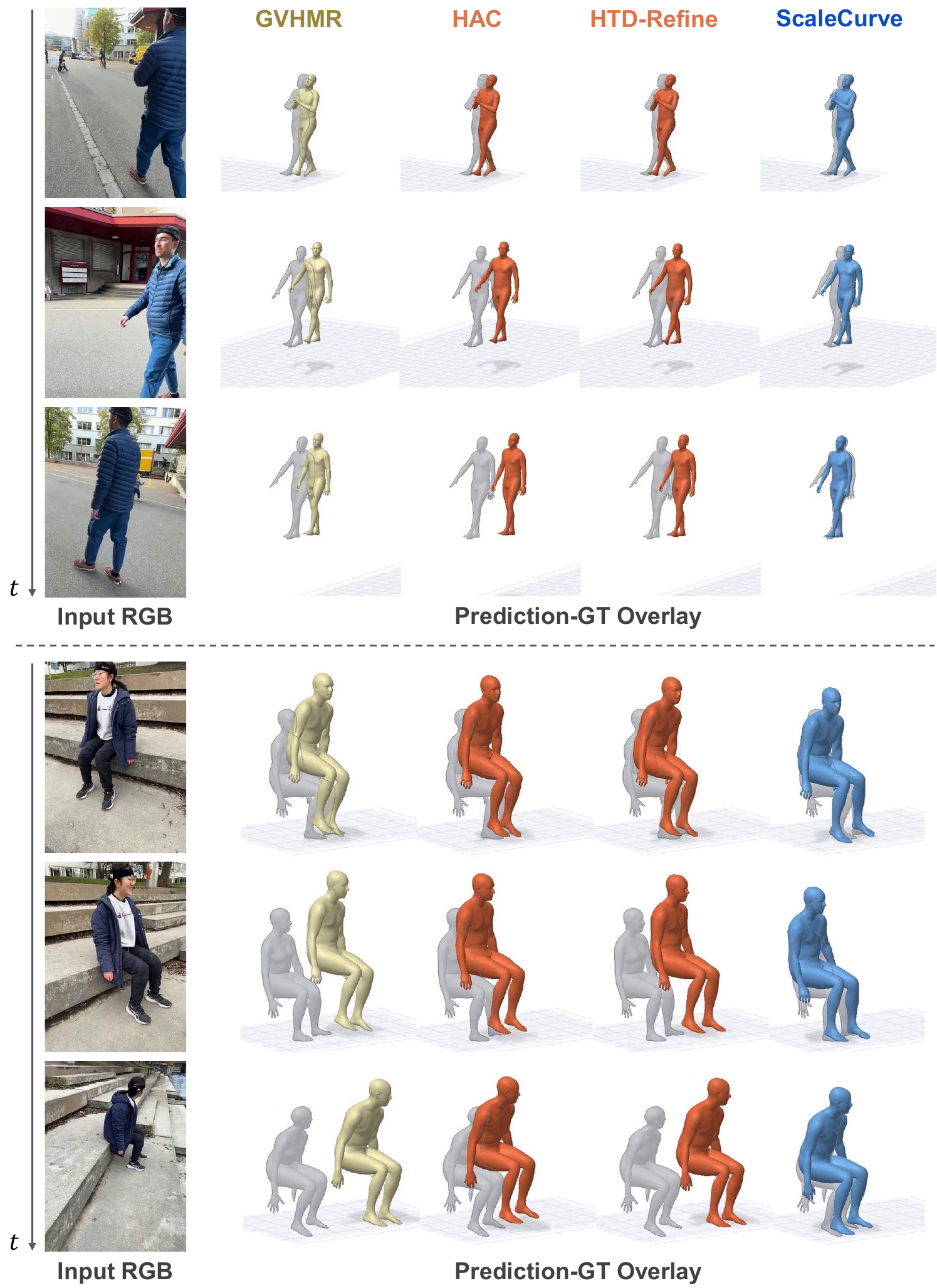}
\caption{\textbf{Qualitative human motion comparisons on EMDB-2.} Three timestamps from each of two supplementary clips, with time progressing downward within each example. GVHMR (yellow), HAC and HTD-Refine (orange), and \method{} (blue) are overlaid with GT (gray) in world coordinates. All methods share a fixed view and display scale within each clip.}
\label{fig:video_mesh_emdb}
\end{figure*}

\begin{figure*}[t]
\centering
\includegraphics[width=.98\textwidth,height=.88\textheight,keepaspectratio]{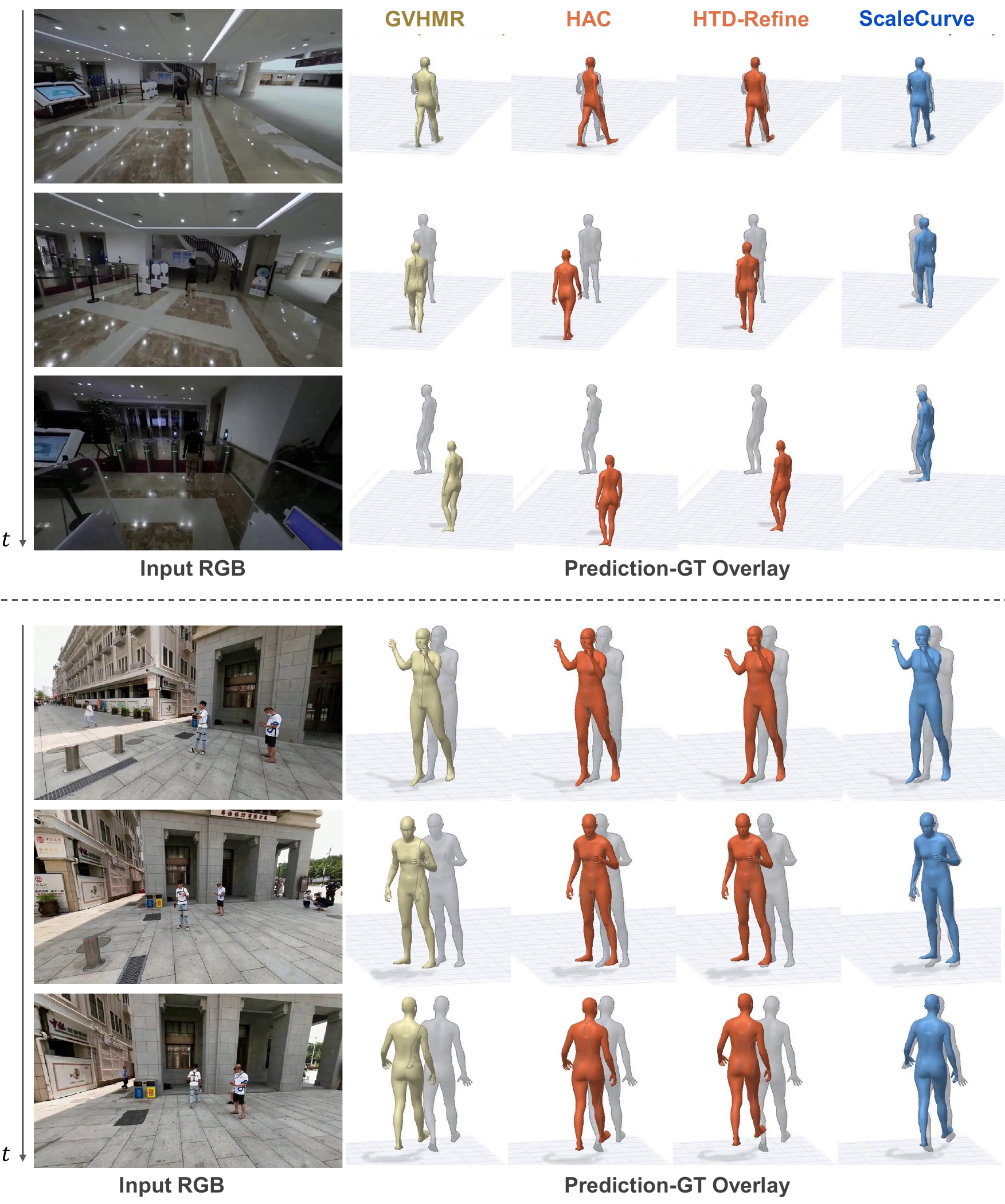}
\caption{\textbf{Qualitative human motion comparisons on SLOPER4D.} Three timestamps from SLOPER4D examples, with time progressing downward within each example. GVHMR (yellow), HAC and HTD-Refine (orange), and \method{} (blue) are overlaid with GT (gray) in world coordinates. All methods share a fixed view and display scale within each clip.}
\label{fig:video_mesh_sloper}
\end{figure*}

\begin{figure*}[t]
\centering
\includegraphics[width=.98\textwidth,height=.88\textheight,keepaspectratio]{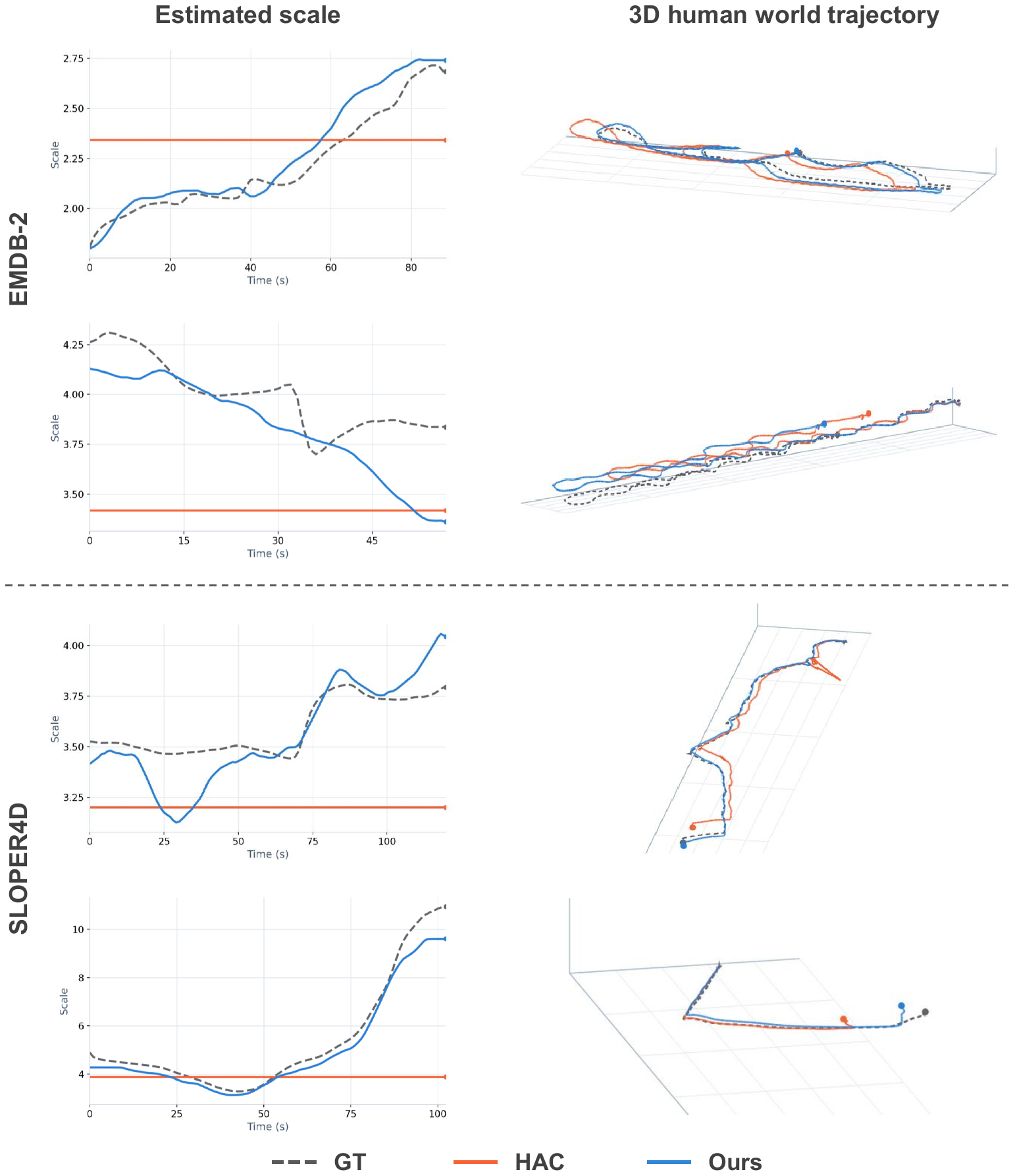}
\caption{\textbf{Qualitative scale and trajectory comparisons.} Estimated scales (left) and world-space human root trajectories (right) for example sequences from EMDB-2 (top) and SLOPER4D (bottom). Gray dashed curves denote the GT-assisted temporal oracle in the scale plots and GT in the trajectory plots.}
\label{fig:video_scale_trajectory}
\end{figure*}

\begin{figure*}[t]
\centering
\includegraphics[width=.8\textwidth]{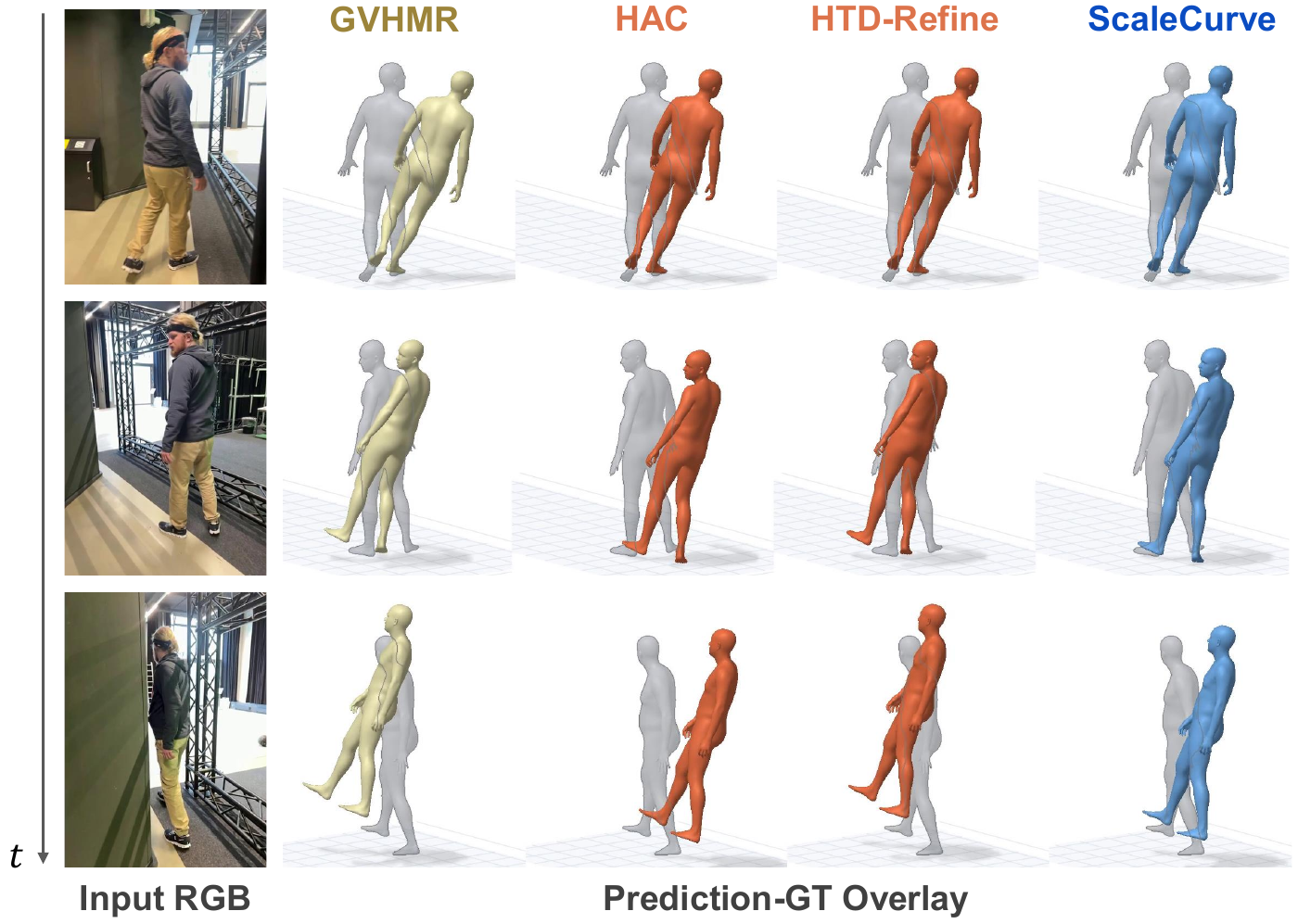}
\caption{\textbf{Failure case: persistent body-orientation errors.} Predictions at three timestamps are overlaid with GT (gray), with time progressing downward. Method colors and the display protocol follow \cref{fig:video_mesh_emdb}.}
\label{fig:video_mesh_failure}
\end{figure*}

\section{Additional Qualitative Results}
\label{sec:qualitative}
We provide additional qualitative comparisons of world-space human motion, estimated scales, and human trajectories, together with a failure example. The supplementary videos included with this submission show these results in motion and provide further examples from both datasets.

\paragraph{World-space human motion.}
\Cref{fig:video_mesh_emdb,fig:video_mesh_sloper} compare GVHMR, HAC, HTD-Refine, and \method{} through GT overlays. On EMDB-2, the baselines develop positional offsets, while \method{} maintains closer alignment over time, including during elevation changes. The SLOPER4D examples show substantial displacement from GT as the camera moves or the subject turns, which our reconstruction reduces. All methods use GVHMR body estimates and share a fixed view and rendering scale within each clip.

\paragraph{Scale and human trajectories.}
\Cref{fig:video_scale_trajectory} shows estimated scales and corresponding human root trajectories for two examples per dataset. Our correction captures temporal changes absent from HAC's constant scale, particularly the pronounced late increase in the second SLOPER4D example. The resulting trajectories follow the GT paths and spatial extent more closely, although local discrepancies remain. The GT-assisted temporal oracle is used only for diagnosis, not by our method.

\paragraph{Failure case.}
\Cref{fig:video_mesh_failure} shows a persistent body-orientation mismatch across all four methods. The reconstructed bodies remain tilted relative to GT across timestamps, even after our calibration and refinement. Temporal scale correction preserves camera rotations and cannot directly resolve orientation errors. Body refinement can adjust orientation, but may retain errors when initialization and motion cues provide insufficient corrective evidence.
\section{Limitations}
\label{sec:limitations}
Despite improved world-space consistency, \method{} depends on usable ground geometry and reliable body-depth estimates. Persistent errors can bias scale observations despite robust temporal optimization. Calibration preserves camera rotations and displacement directions, leaving errors beyond scale uncorrected. Refinement may retain orientation errors when initialization and motion cues are ambiguous, as shown in \cref{fig:video_mesh_failure}. Our evaluation focuses on single-person sequences. Future work could extend the framework to multi-person scenes, incorporate complementary metric cues, and jointly refine camera and body orientation while preserving calibrated metric scale.


\end{document}